\documentclass[letterpaper]{article} 
\usepackage{aaai25}  
\usepackage{times}  
\usepackage{helvet}  
\usepackage{courier}  
\usepackage[hyphens]{url}  
\usepackage{graphicx} 
\usepackage{natbib}  
\usepackage{caption} 
\usepackage{algorithm}
\usepackage{algorithmic}
\usepackage{natbib}
\usepackage{algorithm}
\usepackage{amsmath}
\usepackage{adjustbox}
\usepackage{multirow}
\usepackage{amssymb}
\usepackage{anyfontsize}
\usepackage{amsthm} 

\newtheorem{definition}{Definition} 
\usepackage{blindtext}
\usepackage{multirow}
\usepackage{amssymb}
\usepackage{pifont}
\usepackage{booktabs} 
\usepackage{color, colortbl}
\usepackage[dvipsnames]{xcolor}
\usepackage[dvipsnames]{xcolor}
\usepackage[labelformat=simple]{subcaption}

\usepackage{algorithm}
\usepackage{algorithmic}
\definecolor{Gray}{gray}{0.9}
\definecolor{LightCyan}{rgb}{0.88, 1, 1}

\usepackage{newfloat}
\usepackage{listings}
\DeclareCaptionStyle{ruled}{labelfont=normalfont,labelsep=colon,strut=off} 
\floatstyle{ruled}
\newfloat{listing}{tb}{lst}{}
\floatname{listing}{Listing}
\title{Pre-training Graph Neural Networks on 2D and 3D Molecular Structures \\by using Multi-View Conditional Information Bottleneck}

\author{
    Van Thuy Hoang and O-Joun Lee\thanks{Corresponding author: O-Joun Lee (Tel.: +82-2-2164-5516)}
}
\affiliations{

    Department of Artificial Intelligence, The Catholic University of Korea \\
    {\{hoangvanthuy90, ojlee\}@catholic.ac.kr}
}

\usepackage{bibentry}

\begin{document}

\maketitle

\begin{abstract}

Recent pre-training strategies for molecular graphs have attempted to use 2D and 3D molecular views as both inputs and self-supervised signals, primarily aligning graph-level representations.
However, existing studies remain limited in addressing two main challenges of multi-view molecular learning:
(1) discovering shared information between two views while diminishing view-specific information and
(2) identifying and aligning important substructures, e.g., functional groups, which are crucial for enhancing cross-view consistency and model expressiveness.
To solve these challenges, we propose a Multi-View Conditional Information Bottleneck framework, called MVCIB, for pre-training graph neural networks on 2D and 3D molecular structures in a self-supervised setting.
Our idea is to discover the shared information while minimizing irrelevant features from each view under the MVCIB principle, which uses one view as a contextual condition to guide the representation learning of its counterpart.
To enhance semantic and structural consistency across views, we utilize key substructures, e.g., functional groups and ego-networks, as anchors between the two views.
Then, we propose a cross-attention mechanism that captures fine-grained correlations between the substructures to achieve subgraph alignment across views.
Extensive experiments in four molecular domains demonstrated that MVCIB consistently outperforms baselines in both predictive performance and interpretability.
Moreover, MVCIB achieved the 3d Weisfeiler-Lehman expressiveness power to distinguish not only non-isomorphic graphs but also different 3D geometries that share identical 2D connectivity, such as isomers.

\end{abstract}
%

\begin{figure}[t]
\centering
  \includegraphics[width=  \linewidth]{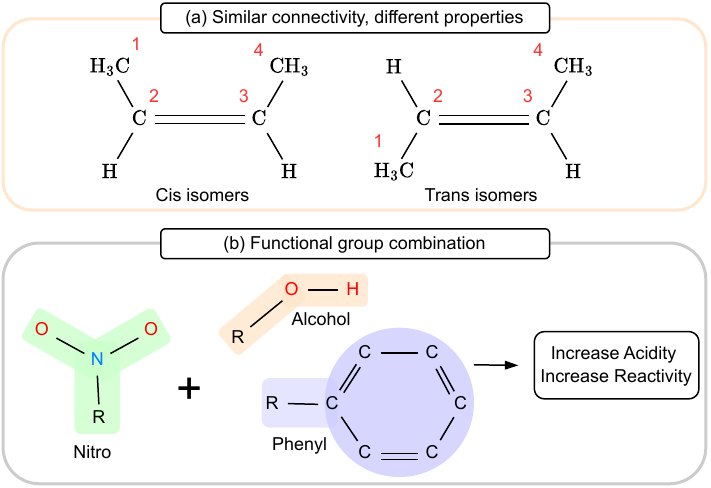}
  \caption{ (a) It is difficult to distinguish cis and trans isomers only based on the 2D molecular graphs.
  (b) The incorporation of a Nitro group (R-${NO}_{2}$) into an Aromatic Hydroxyl group (R-$C_6H_5$, R-$OH$) could change its chemical property.
  }
  \label{fig:problem_1}
\end{figure}

\section{Introduction}



Pre-training Graph Neural Networks (GNNs) for molecules has emerged as a promising approach to address the challenge of limited labels of molecular data \cite{DBLP:conf/nips/RongBXX0HH20,DBLP:conf/nips/LuongS23}.
Most recent methods have mainly focused on 2D molecular graphs, which represent atoms as nodes and chemical bonds as edges.
However, relying only on 2D information can be insufficient to determine chemical properties of molecules \cite{DBLP:conf/iclr/LiuWLLGT22}.
As shown in Fig.~\ref{fig:problem_1} (a), cis and trans isomers have the same 2D connectivity, yet differ in the 3D atom positions, leading to different properties.
That is, cis isomers tend to be more soluble in polar solvents and exhibit a higher boiling point than trans isomers.
However, representations of 2D molecular structures are difficult to reflect such 3D geometric features. 

Recently, 2D and 3D molecular information have been jointly used in the pre-training of GNNs to enhance the quality of representations \cite{DBLP:journals/tmlr/KimNKLAS24,DBLP:conf/icml/StarkBCTDGL22}.
Most strategies aim to leverage the complementary information between the two views by employing contrastive learning objectives to maximize agreement between graph-level representations \cite{DBLP:conf/icml/StarkBCTDGL22,DBLP:conf/iclr/LiuWLLGT22}.
For example, GraphMVP \cite{DBLP:conf/iclr/LiuWLLGT22} introduces a joint training framework that combines contrastive learning with a reconstruction objective to align graph-level representations.
More recently, a few methods have extended contrastive objectives to the subgraph level, aiming to align semantic substructures, e.g., functional groups, to learn chemically informed representations \cite{DBLP:journals/tmlr/KimNKLAS24,DBLP:conf/nips/LuongS23}.
To sum up, most recent studies adopt the contrastive objectives to maximize the agreement between views on graph- or subgraph-level representations.

However, two main challenges limit the recent studies.
\textbf{(1)} Most recent multi-view strategies are limited to extracting shared and task-relevant information while minimizing irrelevant features from each view.
From the multi-view learning perspective, each view is expected to contain a mixture of shared information and view-specific information, as they originate from the same molecule and share the same label \cite{DBLP:conf/iclr/Tsai0SM21}.
Recent studies show that maximizing the shared features is crucial for enhancing the model generalization for downstream tasks \cite{DBLP:conf/iclr/Federici0FKA20}.
However, recent studies focus only on maximizing mutual information between representations from different views, without considering shared and view-specific information. 
\textbf{(2)} The recent studies are limited in their ability to identify and align important substructures across views.
That is, while graph structures can differ between 2D and 3D views, the important substructures, e.g., functional groups, are invariant and commonly play a critical role in determining molecular properties \cite{DBLP:journals/tmlr/KimNKLAS24}.
Through empirical experiments, we showed that capturing and aligning these substructures could enhance the model's expressiveness in distinguishing not only non-isomorphic graphs but also isomers with identical 2D connectivity but differ in 3D geometry (Appendix D.6).
Moreover, the co-occurrences of functional groups can be significant features for property prediction.
For example, Fig.~\ref{fig:problem_1} (b) shows that adding a nitro group (${NO}_{2}$) to a molecule containing a phenyl ($C_6H_5$) and alcohol (${OH}$) can alter the chemical properties.

In this paper, we propose the Multi-View Conditional Information Bottleneck (MVCIB), a GNN pre-training method designed to maximize shared information and align important substructures across molecular views.
%
%
%
MVCIB is characterized by two main contributions.
\textbf{(1)} The key contribution of MVCIB lies in its ability to maximize the shared and task-relevant information while minimizing irrelevant information from specific views in a self-supervised setting.
We address this problem with the Information Bottleneck (IB) principle for learning compressed yet informative representations.
However, applying the IB principle in a self-supervised setting is challenging due to the absence of labels, making it difficult to capture task-relevant information.
To solve it, we propose the MVCIB principle, which utilizes a conditional compression strategy by using one molecular view as a conditional context to guide the representation learning of the other view.
This is based on the multi-view assumption that, although each view could differ in perspective, all views originate from the same underlying molecule.

To align important substructures across views, we first sample key subgraphs from molecular views to serve as anchor points, followed by a cross-attention mechanism to align the representations across views.
\textbf{(2-1)} We extract two types of important subgraphs, i.e., functional groups and ego-networks rooted at each node, to enable MVCIB to capture both chemical semantics and contextual structures, enhancing the model's transferability and expressiveness, respectively.
That is, the model could benefit from learning correct patterns, e.g., functional groups, during the pre-training phase, enhancing its ability to transfer knowledge from large-scale unlabeled datasets to downstream datasets.
To extract functional groups, we employ the BRICS algorithm \cite{degen2008art} to decompose molecules into chemical subgraphs.
%
\textbf{(2-2)} We propose a cross-attention mechanism to align subgraph-level representations across views, where each subgraph representation from one view is used to attend over representations in the other view.
By doing so, the cross-attention scores could enable MVCIB to learn fine-grained cross-view correlations between subgraph representations and align the important substructures.

\section{Related work}

Recent studies have jointly used 2D and 3D molecular information to pre-train GNNs for improving molecular property prediction \cite{DBLP:conf/iclr/LiuWLLGT22,DBLP:conf/kdd/ZhuXW0QZLL22}.
The main idea is to learn representations from two views and then align the graph-level representations under the contrastive objectives \cite{DBLP:conf/iclr/XiaZHG0LLL23,DBLP:conf/icml/StarkBCTDGL22}.
For example, GraphMVP \cite{DBLP:conf/iclr/LiuWLLGT22} employs a joint contrastive learning and reconstruction task to align graph-level representations from two views.
The idea of 3D Infomax \cite{DBLP:conf/icml/StarkBCTDGL22} is to infer the molecular geometry from 2D molecular graphs by maximizing the agreement between two view representations.
However, most recent strategies overlook discovering shared information and do not identify functional groups, which are crucial for determining property prediction.
Besides, several augmentation-based strategies extend 2D molecular learning to a multi-view problem by using augmentation schemes, e.g., edge perturbation, with contrastive objectives \cite{DBLP:conf/iclr/LiuWLLGT22,DBLP:conf/kdd/QiuCDZYDWT20}.
However, these studies do not learn the complementary information from 2D and 3D molecular structures, limiting their ability to capture the 3D geometric features.
In contrast, MVCIB is designed to preserve and align chemical substructures between views, followed by maximizing the shared information and minimizing irrelevant features across views.

Recently, important molecular substructures, e.g., functional groups, have been incorporated into pre-training methods to improve knowledge transfer to downstream tasks \cite{DBLP:conf/nips/ZhangLWLL21,DBLP:conf/nips/LiuSZZKWC23}.
These substructures are commonly determined by using chemical rules or manual annotation \cite{inae2024motifaware}.
For example, GraphFP \cite{DBLP:conf/nips/LuongS23} decomposes 2D molecular structures into a bag of fragments based on a subgraph dictionary, constructed from frequently occurring substructures across molecular datasets.
However, these strategies primarily focus on 2D molecular structures and are thus limited in capturing the 3D geometric features.
More recently, several studies \cite{DBLP:journals/tmlr/KimNKLAS24,DBLP:journals/bioinformatics/HuangFSW024} have proposed using both 2D and 3D molecular views to effectively encode the chemical substructures into representations.
For instance, HoliMol \cite{DBLP:journals/tmlr/KimNKLAS24} extracts important substructures from both views by decomposing molecules into fragments and then aligns the graph-level representations using a contrastive objective.
However, these methods rely only on contrasting graph-level representations without considering the shared information between two views, which could retrain noisy information and degrade generalization performance across downstream tasks \cite{DBLP:conf/iclr/Federici0FKA20}.
In contrast, MVCIB not only focuses on subgraph-level alignment but also maximizes the shared information between molecular views and minimizes the irrelevant information.

\section{Problem Descriptions}

We study the problem of multi-view molecular learning under the MVCIB principle.
Thus, we first introduce the notation, followed by a definition of MVCIB, which extends the IB principle to the multi-view setting.

Let ${G}_{2D} = ( {X}, {E}_{2D})$ denote a 2D molecular graph with initial atom features ${X}$ and the set of edges ${E}_{2D}$. 
Let ${G}_{3D}= ( {X}, {E}_{3D}, {R})$ denote the corresponding 3D molecular graph, where ${E}_{3D}$ denotes the set of edges, and ${R} \in \mathbb{R}^{N \times 3}$ contains the 3D coordinates of the $N$ atoms.

Recently, the Information Bottleneck (IB) principle \cite{DBLP:journals/corr/physics-0004057} has been applied to graphs to learn compressed yet informative representations.
\begin{definition}[IB]
\label{def:IB}
Given an input graph $G$ and its label $Y$, the IB principle learns the compressed yet sufficient representation $\textbf{Z}$ by optimizing the following objective: 
\begin{eqnarray}
\label{eq:IB}
\underset{\textbf{Z}} {\mathop{\min}} \  I(G,\textbf{Z})  - \beta I(\textbf{Z},Y),
\end{eqnarray}
where the first term encourages compression of the input, the second term is for predictive sufficiency, and $\beta$ is a hyperparameter controlling the balance between the two terms.
\end{definition}


In the context of self-supervised learning, our goal is to optimize the first term in Eq. \ref{eq:IB}, omitting the use of label $Y$.
From a multi-view perspective, let $G_{2D}$ and $G_{3D}$ denote the 2D and 3D views from the same molecule, respectively.
Each view can contain view-specific information, which can be considered as noise from its counterpart view.
However, as both views originate from the same molecule and share identical labels, their representations need to encode the shared features that are most relevant for downstream tasks.
Here, we present the formulation for the 2D view only, as the formulation for the 3D view follows analogously.
From the 2D view, we aim to compress $G_{2D}$ into a compressed yet shared representation $\textbf{Z}^{CIB }_{2D}$ conditioned by the $G_{3D}$.
\begin{definition}[MVCIB]
\label{def:CIB}
When $G_{3D}$ is already observed, the mutual information $I({G}_{2D} | {\textbf{Z}}^{CIB}_{2D})$ can be factorized to reveal shared information in the 2D view.
By conditioning on $G_{3D}$, the shared information can be discovered as:
\begin{equation}
\label{eq:def_cib} 
\resizebox{0.47\textwidth}{!}{$
\textbf{Z}^{CIB }_{2D} = \underset{\textbf{Z}^{CIB}_{2D}} {\mathop{\arg \min }}\,  \underbrace{I\left( {G_{2D}};{\textbf{Z}}^{CIB}_{2D}|{G_{3D}} \right)}_{\text{Conditional compression}} -\underbrace{  \beta  I\left( {\textbf{Z}}^{CIB}_{2D} ;{G_{3D}} \right)}_{\text{Sufficient prediction}} . $} \     
\end{equation}
\end{definition}
The second term, $I({\textbf{Z}}^{CIB}_{2D}; G_{3D})$, aims to maximize sufficiency, ensuring that the representation ${\textbf{Z}}^{CIB}_{2D}$ retains as much shared information as possible from the complementary view $G_{3D}$, thereby increasing its sufficiency for label prediction.
Thus, the mutual information $I({G}_{2D} | {\textbf{Z}}^{CIB}_{2D})$ can be reduced by minimizing the first term $I({G_{2D}};{\textbf{Z}}^{CIB}_{2D}|{G_{3D}})$.
This penalizes the information in ${\textbf{Z}}^{CIB}_{2D}$ that is unique to $G_{2D}$ and not relevant to the $G_{3D}$.
Similarly, from 3D view, we can exploit a shared information $\textbf{Z}^{CIB }_{3D}$ when $G_{2D}$ is already observed.


\section{Methodology}

\begin{figure*}[t]
\centering
\includegraphics[width=  \linewidth]{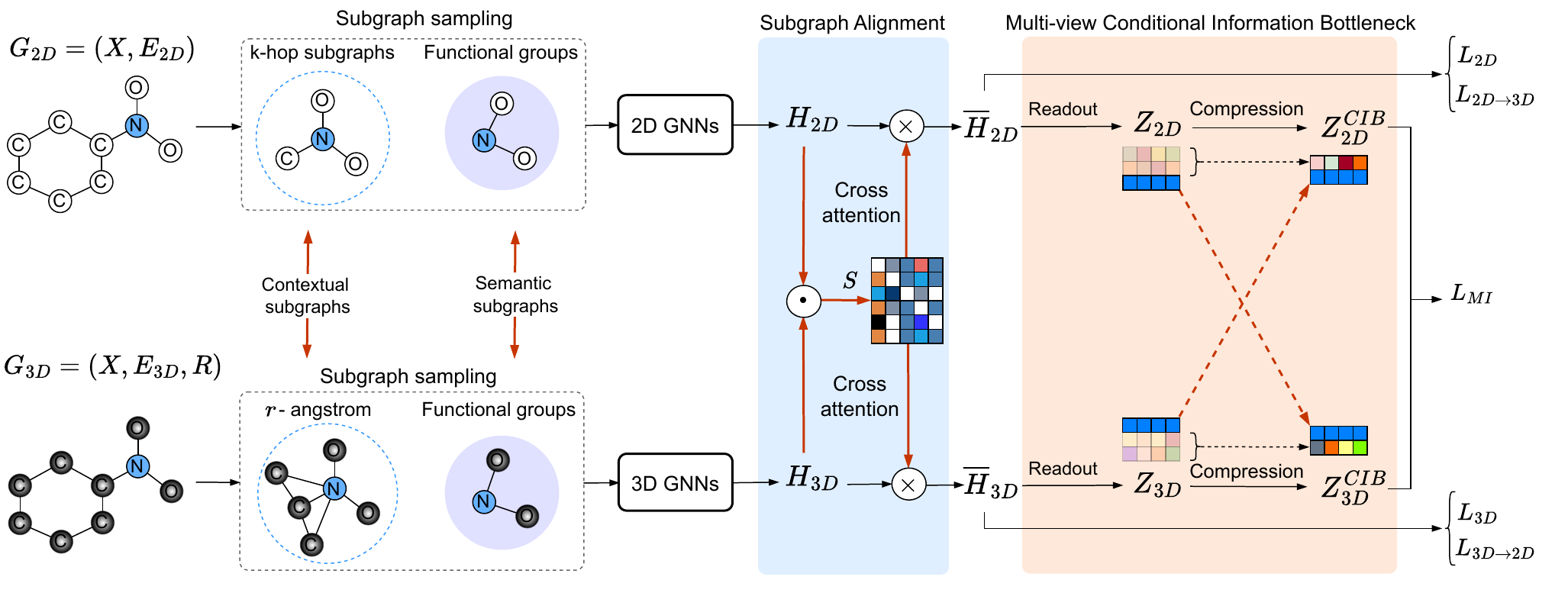}
\caption{An overall architecture of MVCIB, consisting of (\textbf{i} \& \textbf{ii}) subgraph sampling and alignment and (\textbf{iii}) MVCIB principle.}
\label{fig:architecture}
\end{figure*}


\subsection{Multi-View Subgraph Sampling}
Fig.~\ref{fig:architecture} presents the architecture of MVCIB.
For subgraph sampling, we simultaneously sample the two types of subgraphs, i.e., contextual and semantic subgraphs, for each node within both the 2D and 3D molecular graphs.
To sample contextual subgraphs in the 2D graphs, for each given node $v$, we extract $k$-hop ego-network $G_{2D}[N_{k}{(v)}]$ rooted at $v$, where $N_{k}{(v)}$ denotes the set of neighbor nodes within $k$-hop distance.
In contrast, for the 3D graphs, we sample contextual subgraphs $G_{3D} [N_{r}{(v)}]$ containing atoms that lie within an $r$-\text{\AA} radius of node $v$.

To extract semantic subgraphs, i.e., functional groups, we employ the BRICS algorithm \cite{degen2008art}, which incorporates domain knowledge to decompose molecules into chemical substructures.
Formally, given a 2D molecular graph $G_{2D}$, we define the set of functional group fragments as $F = \{ G^{(0)},G^{(1)}, \cdots, G^{(m)} \}$, where $G^{(i)} = (V^{(i)}, E^{(i)})$ denotes the $i$-th subgraph,
$m$ is the number of fragments,
and $V^{(i)}\cap V^{(j)} = \{\emptyset \}$.
For the 3D graph, we extract the same functional groups as the 2D graph, as they are structurally invariant and can be consistently recognized across views.


\subsection{Multi-View Subgraph Alignment}

\subsubsection{Subgraph Learning}
For 2D molecular graphs, we adopt the GIN encoder \cite{DBLP:conf/iclr/XuHLJ19}, which is a powerful GNN model with high structure distinguishability.
Specifically, given a target node $v$ in $G_{2D}$, we obtain two types of representations, i.e., contextual and semantic representations.
To learn the contextual representation of a node $v$, we first learn the representation of any node $u$ within the ego-network $G_{2D}[N_{k}{(v)}]$ rooted at $v$.
Then, the contextual representation of node $v$ is obtained via a readout function, as:
\begin{eqnarray}
 \textbf{H}_{2D}^{u  } & = & \textbf{GNN} \left( G_{2D} \left[ {{N}_{k}}\left( v \right) \right]  \ | \ u \in N_k\left(v \right) \right) , \\
 \textbf{H}_{2D}^{v | \text{Context}} & = & \text{Readout} \left( \textbf{H}_{2D}^{u  }  \ | \ u \in {{N}_{k}} \left( v \right) \right) , \
\end{eqnarray}
where $\textbf{GNN} (\cdot)$ is the GIN encoder and $\text{Readout}(\cdot)$ is a summation function.
Similarly, to obtain the semantic representations of node $v$, we apply the GIN encoder on its functional group subgraph $G^{(i)}$, as follows:
\begin{eqnarray}
 \textbf{H}_{2D}^{u  } & = & \textbf{GNN} \left(G^{(i)}  \ | \ u \in G^{(i)} \right) , \\
 \textbf{H}_{2D}^{v | \text{Semantics}} & = & \text{Readout} \left( \textbf{H}_{2D}^{u  }  \ | \ u \in G^{(i)} \right) . \
\end{eqnarray}
By doing so, our encoder can capture the contextual and semantic subgraph-level information rooted at each node $v$.
We then incorporate the representations of the node $v$, as: 
\begin{eqnarray}
 \textbf{H}_{2D}^{v} & = & \textbf{FUSE} \left( \textbf{H}_{2D}^{v| \text{Context}} , \textbf{H}_{2D}^{v | \text{Semantics}}\right) ,\
\end{eqnarray}
where \textbf{FUSE}($\cdot$) is concatenation.

For the 3D molecular graph, we employ an Equivariant Graph Neural Network (EGNN) encoder \cite{DBLP:conf/icml/SatorrasHW21}, a powerful architecture designed to preserve equivariance to geometric transformations, e.g., rotations. 
Analogously, for the node $v$, we apply EGNN to obtain the contextual representation $\textbf{H}_{3D}^{v | \text{Context}}$ in $G_{3D}[N_{r}{(v)}]$ and semantic representation $\textbf{H}_{3D}^{v| \text{Semantics}}$ in $G^{(i)}$.
Then, the subgraph-level representation of a node $v$ from the 3D view can be defined as: 
\begin{eqnarray}
 \textbf{H}_{3D}^{v} & = & \textbf{FUSE} \left( \textbf{H}_{3D}^{v| \text{Context}} , \textbf{H}_{3D}^{v | \text{Semantics}}\right) .\
\end{eqnarray}

\subsubsection{Subgraph Alignment}


%
We now align important substructures by using a cross-attention mechanism, where each atom representation in one view is used to attend over representations in the other view.
Specifically, we compute an affinity matrix $\textbf{S}$ to quantify how similar the representations of each atom are across the different views, as:
\begin{eqnarray}
\textbf{S} = \left( {{{W}}_{2D}}{{\textbf{H}}_{2D}} \right){{\left( {{{W}}_{3D}}{{\textbf{H}}_{3D}} \right)}^{\intercal}}, \
\end{eqnarray}
where ${{W}}_{2D}$ and ${{W}}_{3D}$ are learnable matrices and $\textbf{S}_{ij}$ denotes the similarity of the $i$-th atom from $\textbf{H}_{2D}$ and the $j$-th atom from $\textbf{H}_{3D}$.
For 2D-queried cross-attention, we employ a cross-attention mechanism, where each 2D representation serves as a query to compute attention weights over the 3D representations. Then, we construct an attended 3D representation $\overline{\textbf{H}}_{3D}^{i}$ based on given 2D representation ${\textbf{H}}^i_{2D}$ by applying a weighted combination over 3D representations:
\begin{eqnarray}
{\overline{\textbf{H}}_{3D}^{i}}=\sum\nolimits_{j=1}^{N}{{{\xi}_{ij}}\textbf{H}_{3D}^{j}}\text{ , } \ \ {{\xi }_{ij}}=\frac{\exp \left(  {{\textbf{S}}_{ij}} \right)}{\sum\nolimits_{j=1}^{N}{\exp \left(  {{\textbf{S}}_{ij}} \right)}} \ .
\end{eqnarray}

Likewise, we can compute the 2D representations that are attended by 3D representations by considering each 3D representation as a query:
\begin{eqnarray}
\overline{\textbf{H}}_{2D}^{j}=\sum\nolimits_{i=1}^{N}{{{\zeta }_{ij}}\textbf{H}_{2D}^{i}}\text{ , } \ \ {{\zeta }_{ij}}=\frac{\exp \left(   {{\textbf{S}}_{ij}} \right)}{\sum\nolimits_{i=1}^{N}{\exp \left(   {{\textbf{S}}_{ij}} \right)}}  \ .
\end{eqnarray}


\subsection{Multi-View Conditional Information Bottleneck}
\label{sect:graph_level_align}



We now present our MVCIB principle by maximizing the shared information across views in an SSL setting.
Specifically, from the 2D view, we compress $G_{2D}$ into shared information $\textbf{Z}^{CIB}_{2D}$ and discard the noise from $G_{2D}$ that is not relevant to the 3D view.
Based on Definition \ref{def:CIB}, we can minimize $I( G_{2D} | \textbf{Z}^{CIB}_{2D} )$ under MVCIB principle, as:
\begin{eqnarray}
\label{eq:2d1}
\resizebox{0.44\textwidth}{!}{$
  I\left( G_{2D} | \textbf{Z}^{CIB}_{2D} \right) = {I\left( {G_{2D}};\textbf{Z}_{2D}^{CIB}| G_{3D} \right)} - \beta_{1} I\left( \textbf{Z}_{2D}^{CIB};{G_{3D}} \right) \ . $} 
\end{eqnarray}
Similarly, from the 3D view, we can minimize the mutual information $I( {G}_{3D} | \textbf{Z}^{CIB}_{3D} )$ as:
\begin{eqnarray}
\label{eq:3d1}
\resizebox{0.44\textwidth}{!}{$
  I\left( G_{3D} | \textbf{Z}^{CIB}_{3D} \right) = {I\left( {G_{3D}};\textbf{Z}_{3D}^{CIB}|{G_{2D}} \right)} - \beta_{2} I\left( \textbf{Z}_{3D}^{CIB};{G_{2D}} \right) \ .  $}
\end{eqnarray}

We now jointly minimize Eq. \ref{eq:2d1} and Eq. \ref{eq:3d1}.
By combining and modifying the Lagrangian multipliers $\beta_1$ and $\beta_2$, we propose to minimize the loss function $L_{MI}$, as:
\begin{eqnarray}
\label{eq:cib}
L_{MI} &= \left[  I\left( {G_{2D}};\textbf{Z}_{2D}^{CIB}|{G_{3D}} \right) + {I\left( {G_{3D}};\textbf{Z}_{3D}^{CIB}|{G_{2D}} \right)} \right]  \nonumber \\ 
& -\alpha  \left[  I\left( \textbf{Z}_{2D}^{CIB};{G_{3D}} \right) + I\left(\textbf{Z}_{3D}^{CIB};{G_{2D}} \right)\right] , 
\end{eqnarray}
where $\alpha$ controls the trade-off between the two terms. 
By considering $\textbf{Z}_{2D}^{CIB}$ and $\textbf{Z}_{3D}^{CIB}$ into a shared latent space, we can derive an upper bound on $L_{MI}$, as:
\begin{eqnarray}
\label{eq:final}
& L_{MI} \le D_{{SKL}}\left( p\left( \textbf{Z}_{2D}^{CIB}|{G_{2D}} \right)\,\|\, p\left( \textbf{Z}_{3D}^{CIB}|{G_{3D}} \right) \right)  \nonumber \\ 
&- \alpha I\left(  \textbf{Z}_{2D}^{CIB};  \textbf{Z}_{3D}^{CIB} \right) \ ,
\end{eqnarray}
where $D_{{SKL}}( \cdot \,\|\, \cdot )$ denotes a symmetrized KL divergence between two distributions.
The proof of Eq. \ref{eq:final} is provided in Appendix A.1. 
For the first term, we can directly compute the average of the expected KL divergences between distributions ${{D}_{KL}}( p( \textbf{Z}_{2D}^{CIB}|{{\textbf{Z}}_{2D}} )||p( \textbf{Z}_{3D}^{CIB}|{{\textbf{Z}}_{3D}} ) )$ and ${{D}_{KL}}( p( \textbf{Z}_{3D}^{CIB}|{{\textbf{Z}}_{3D}} ) || p( \textbf{Z}_{2D}^{CIB}|{{\textbf{Z}}_{2D}} ) )$, that both distributions have known densities.
To optimize the second term, i.e., $-I(\textbf{Z}_{2D}^{CIB};  \textbf{Z}_{3D}^{CIB} )$, we adopt the Jensen–Shannon estimator \cite{DBLP:conf/iclr/HjelmFLGBTB19}, which can enable a tractable approximation of mutual information via neural networks.

The final representation is obtained by concatenating $\textbf{Z}_{2D}^{CIB}$ and $\textbf{Z}_{3D}^{CIB}$, which is then used for downstream tasks.

\subsection{Multi-Task Objectives}
 
\subsubsection{2D-View SSL}

From the 2D view, we employ a graph reconstruction loss $L_{2D}$ to encourage MVCIB to preserve the 2D graph structure.
Specifically, given the representation matrix $\overline{H}_{2D}$, we compute the pairwise similarity between any two nodes $v_i$ and $v_j$ by using cosine similarity, i.e., ${P_{i,j}}=\frac{\overline{H}^{{i} }_{2D}{\overline{H}^{{j}}_{2D}}}{\| {\overline{H}^{{i}}_{2D}} \|\| {\overline{H}^{{j}}_{2D}} \|}$, \cite{zhang2020graph}. 
Then, the reconstruction loss can be defined as follows:
\begin{eqnarray}
& L_{2D}=\frac{1}{{{\left| V \right|}^{2}}}\| A- P \|_{F}^{2} \ ,   
\end{eqnarray}
where $A$ denotes the original adjacency matrix of $G_{2D}$.

\subsubsection{3D-View SSL}
We adopt a 3D position denoising as a SSL task for 3D graph recovery, following \citet{DBLP:conf/iclr/GodwinSGSRVKB22,DBLP:conf/iclr/ZaidiSMKTSBPG23}.
Specifically, Gaussian noise is added to the input positions of atoms, and the model is trained to predict the noise from the perturbed input.
Let $\textbf{R} = \{r_1, r_2, \cdots, r_n \}$, where $r_i$ denotes the position of the $i$-th atom, and the noisy version of the atom positions is denoted by:
$\hat{\textbf{R}} = \{r_1 + \sigma \epsilon_1, r_2+ \sigma \epsilon_2, \cdots, r_n+ \sigma \epsilon_n \}$, where $\sigma$ is the scaling factor and $\epsilon \in \mathcal{N}(\textbf{0},\textbf{I})$.
The denoising loss is calculated by the cosine similarity between the predicted and ground-truth noises as:
\begin{eqnarray}
{{L}_{3D}}=\frac{1}{B}\sum\limits_{i=1}^{B}{\sum\limits_{j=1}^{N} \left[ {1-\cos \left( \varepsilon _{j}^{i},\hat{\varepsilon }_{j}^{i} \right)} \right] }, \ 
\end{eqnarray}
where $B$ denotes the number of graphs in a mini-batch,
and $\hat{\epsilon}$ is a prediction output of the model, following the work \cite{abs-2203-04810,LuoCXZL0H23}.

\subsubsection{Cross-View SSL}

To better ensure that the shared information retains view-specific information from both 2D and 3D molecular views, we introduce a cross-view SSL task to align the representations across views.
Specifically, given 2D representations $Z_{2D}^{CIB}$, our goal is to infer geometric distance between atoms, without relying on 3D coordinates.
The spatial reconstruction loss can be defined as:
\begin{eqnarray}
{{L}_{2D\to 3D}}=\frac{1}{N}\sum\nolimits_{{{v}_{i}},{{v}_{j}}}^{{}}{\left\| f\left( {\overline{\textbf{H}}^{{i}}_{2D}} - {\overline{\textbf{H}}^{{j}}_{2D}} \right)-{{D}_{ij}} \right\|} \ , 
\end{eqnarray}
where $f(\cdot)$ denotes a MLP function and $D$ denotes the distance matrix between atoms $v_i$ and $v_j$.

Analogously, given the 3D representation $\overline{H}_{3D} $, we aim to infer the 2D bonding connectivity.
Thus, we compute a 2D connectivity loss from the 3D representation, as:
\begin{eqnarray}
  & {{L}_{3D\to 2D}} = \frac{1}{{{\left| V \right|}^{2}}}\| A- Q \|_{F}^{2}  , \
\end{eqnarray}
where $Q$ is the cosine pairwise similarity matrix of any node pairs in $\overline{H}_{3D}$.
The total loss for the pre-training is as:
\begin{eqnarray}
  L =   L_{MI}  + \beta \left( {{L}_{2D}}+{{L}_{3D}} + {{L}_{2D\to 3D}}+{{L}_{3D\to 2D}} \right),
\end{eqnarray}
where $\beta$ controls the trade-off between the compression and reconstruction terms.



\begin{table*}[tb]
\centering
\renewcommand*{\arraystretch}{0.80}
\setlength{\tabcolsep}{4 pt}
\fontsize{9 pt}{9 pt}\selectfont
\caption{A performance comparison on classification tasks (ROC-AUC).
The top two are highlighted by \textbf{first} and \underline{second}.}

\begin{tabular}{l cc cc cc c c }
\toprule
Methods
&BBBP
&Tox21
&ToxCast 
&SIDER
&ClinTox
&MUV
&HIV 
&BACE 
\\ 
\midrule



Infomax 
&68.39$\pm$0.64 
&72.66$\pm$0.16 
&62.76$\pm$0.54 
&59.02$\pm$0.56 
&58.62$\pm$0.83 
&72.14$\pm$1.25 
&73.55$\pm$0.47 
&77.80$\pm$0.46 

\\
JOAO 
&71.63$\pm$1.11
&73.67$\pm$1.06
&63.30$\pm$0.27
&{63.55$\pm$0.81}
&77.02$\pm$1.64
&69.81$\pm$0.26
&77.55$\pm$1.94 
&74.94$\pm$1.35 

\\

GraphCL 
&68.39$\pm$0.64 
&73.26$\pm$0.59 
&62.76$\pm$0.54 
&61.83$\pm$0.60 
&61.62$\pm$1.25 
&72.14$\pm$1.25 
&73.55$\pm$0.47 
&77.80$\pm$0.46 

\\
GraphLoG 
&66.75$\pm$0.32
&71.64$\pm$0.49 
&61.53$\pm$0.35 
&59.09$\pm$0.53 
&53.76$\pm$0.95 
&72.52$\pm$2.02 
&73.76$\pm$0.29 
&76.60$\pm$1.04 

\\
\midrule
GraphFP 
&72.05$\pm$1.17
&{77.35$\pm$1.40}
&{69.15$\pm$1.92} 
&\underline{65.93$\pm$3.09}
&76.80$\pm$1.83
&71.82$\pm$1.33
&75.71$\pm$1.39 
&80.28$\pm$3.06 

\\
MICRO-Graph
&67.21$\pm$1.85
&71.79$\pm$1.70
&60.80$\pm$1.15 
&60.34$\pm$0.96
&{77.56$\pm$1.56}
&70.46$\pm$1.62
&76.73$\pm$1.07 
&63.57$\pm$1.55 
\\ 

MGSSL
&79.52$\pm$1.98
&74.82$\pm$1.60
&63.86$\pm$1.57 
&57.46$\pm$1.45
&75.84$\pm$1.82
&{73.44$\pm$3.47}
&77.45$\pm$2.94 
&{82.03$\pm$3.79}

\\ 
GROVE 
&{87.15$\pm$0.06}
&68.59$\pm$0.24
&64.45$\pm$0.14 
&57.53$\pm$0.23
&72.53$\pm$0.14
&67.67$\pm$0.12
&75.04$\pm$0.13 
&81.13$\pm$0.14 

\\ 

S-CGIB
&\underline{88.75$\pm$0.49}
&\underline{80.94$\pm$0.17}
&\underline{70.95$\pm$0.27} 
&{64.03$\pm$1.04} 
&\underline{78.58$\pm$2.01}
&\underline{77.71$\pm$1.19}  
&\underline{78.33$\pm$1.34}
&\underline{86.46$\pm$0.81}

\\

\midrule

GraphMVP
&68.78$\pm$0.24
&74.39$\pm$0.37
&62.65$\pm$0.73
&61.32$\pm$1.41
&73.87$\pm$1.22
&72.83$\pm$0.34
&76.58$\pm$0.64
&76.71$\pm$0.92
\\
GraphMVP-G
&70.76$\pm$0.58
&74.71$\pm$0.29
&61.53$\pm$0.78
&59.59$\pm$1.28
&74.69$\pm$1.82
&72.40$\pm$0.19
&76.34$\pm$0.48
&77.03$\pm$0.81

\\
GraphMVP-C
&70.63$\pm$0.78
&74.46$\pm$0.39
&62.59$\pm$0.49
&63.15$\pm$1.19
&75.68$\pm$2.47
&74.71$\pm$1.27
&75.14$\pm$1.06
&80.99$\pm$0.46
\\
Holi-Mol
&70.37$\pm$0.35
&75.11$\pm$0.32
&61.19$\pm$0.38
&61.37$\pm$0.64
&77.81$\pm$0.67
&75.96$\pm$0.56
&76.73$\pm$0.29
&78.82$\pm$0.94
\\
3D-InfoMax
&68.39$\pm$1.35
&73.02$\pm$1.09
&60.73$\pm$0.58
&56.06$\pm$1.33
&60.74$\pm$1.43
&75.31$\pm$1.62
&74.25$\pm$0.95
&78.21$\pm$1.77
\\
\midrule

MVCIB (Ours)
&\textbf{90.76$\pm$0.45}
&\textbf{81.03$\pm$0.47}
&\textbf{72.28$\pm$0.31}
&\textbf{67.34$\pm$0.31}
&\textbf{81.93$\pm$1.06}
&\textbf{77.92$\pm$0.35}
&\textbf{79.75$\pm$0.51}
&\textbf{87.39$\pm$0.21}
\\

\bottomrule
\end{tabular}
\label{tab:acc1}
\end{table*}

\section{Evaluation}

\subsection{Experimental Settings}

\subsubsection{Datasets}

We conducted experiments across four different chemical domains:  
Physiology (BBBP, Tox21, ToxCast, SIDER, ClinTox, and MUV),
Physical Chemistry (ESOL, FreeSolv, and Lipo),
Biophysics (mol-HIV and BACE \cite{wu2018moleculenet}),
and Quantum Mechanics QM9 \cite{DBLP:conf/nips/HuFRNDL21}.
For the pre-training dataset, we considered 456k unlabeled molecules from the ChEMBL database \cite{mayr2018large}.
For the model explainability, we used three datasets, i.e., Benzene, Alkane Carbonyl, and Fluoride Carbonyl, with the presence of ground truth \cite{agarwal2023evaluating}.
We evaluated the expressiveness of MVCIB on graph isomorphism testing datasets \cite{DBLP:conf/icml/BalcilarHGVAH21}.
We randomly split the datasets into training/validation/test sets with a ratio of 6:2:2. 
Dataset statistics are described in Appendix B. 

\subsubsection{Baselines and Implementation Details}
We considered three baseline groups.
The contrastive learning methods are 
Infomax \cite{velickovic2019deep}, 
JOAO \cite{DBLP:conf/icml/YouCSW21}, 
GraphCL \cite{DBLP:conf/nips/YouCSCWS20}, 
and GraphLoG \cite{DBLP:conf/icml/XuWNGT21}.
The subgraph-based methods are 
MICRO-Graph \cite{DBLP:conf/aaai/Subramonian21}, 
MGSSL \cite{DBLP:conf/nips/ZhangLWLL21}, 
GraphFP \cite{DBLP:conf/nips/LuongS23}, 
GROVE \cite{DBLP:conf/nips/RongBXX0HH20}, 
The multi-view methods are
GraphMVP \cite{DBLP:conf/iclr/LiuWLLGT22},
Holi-Mol \cite{DBLP:journals/tmlr/KimNKLAS24}, and
3D-InfoMax \cite{DBLP:conf/icml/StarkBCTDGL22}.
We deliver an open-source implementation of MVCIB for the experiment reproductions\footnote{https://github.com/NSLab-CUK/MVCIB}.
The implementation details are presented in Appendix C. 

\subsection{Performance Analysis}

\subsubsection{Molecular Property Classification Tasks}

\begin{table*}[t]
\caption{A performance comparison on 3D geometric property regression tasks in QM9 dataset in terms of MAE.}
\begin{adjustbox}{width= \linewidth}
\renewcommand*{\arraystretch}{0.80}
\setlength{\tabcolsep}{2 pt}
\fontsize{9 pt}{9 pt}\selectfont
\centering
\begin{tabular}{l ccc ccc c}
\toprule

Methods
&$\mu$
&$\alpha$  
&HOMO
&LUMO
&$\varepsilon_{gap}$
&$R^2$
&zpve

\\
\midrule



Infomax 
&21.6427$\pm$0.2713
&5.8245$\pm$0.0497
&4.6103$\pm$0.0257
&22.5401$\pm$0.1792
&14.4615$\pm$0.1526
&99.6702$\pm$4.8413
&28.8924$\pm$0.2435
\\
JOAO
&13.3878$\pm$0.1478
&19.2648$\pm$0.1783
&10.7519$\pm$0.1024
&18.9824$\pm$1.1556
&21.2218$\pm$1.0699
&101.0046$\pm$5.8518
&25.2529$\pm$0.9581
\\
GraphCL
&4.4489$\pm$0.0822
&15.5298$\pm$1.0676
&7.7433$\pm$0.9018
&18.8141$\pm$0.1287
&6.4682$\pm$0.0275
&127.5179$\pm$6.4723
&10.8973$\pm$0.0285
\\
GraphLoG
&5.7881$\pm$0.0614
&17.2818$\pm$0.1397
&9.3541$\pm$0.6121
&15.8762$\pm$0.6982
&10.5164$\pm$0.9695
&97.8614$\pm$4.3656
&12.8907$\pm$0.9833
\\
\midrule

GraphFP
&0.5838$\pm$0.0174
&1.2528$\pm$0.0725
&0.0074$\pm$0.0008
&0.0071$\pm$0.0003
&0.0087$\pm$0.0005
&53.2751$\pm$2.9725
&0.0043$\pm$0.0001
\\

MICRO-Graph
&0.5241$\pm$0.0073
&1.2649$\pm$0.0114
&0.0048$\pm$0.0009
&0.0085$\pm$0.0004
&0.0079$\pm$0.0003
&44.1453$\pm$2.9864
&0.0034$\pm$0.0005
\\
MGSSL
&0.5746$\pm$0.0059
&1.0613$\pm$0.0128
&\underline{0.0046$\pm$0.0002}
&\underline{0.0047$\pm$0.0003}
&\underline{0.0066$\pm$0.0008}
&38.8678$\pm$1.7486
&0.0083$\pm$0.0006
\\

GROVE
&0.6113$\pm$0.0037
&\underline{1.0296$\pm$0.0124}
&0.0059$\pm$0.0007
&0.0051$\pm$0.0004
&0.0082$\pm$0.0008
&41.9575$\pm$2.0181
&0.0071$\pm$0.0002
\\


S-CGIB
&0.5402$\pm$0.0065
&3.0022$\pm$0.0081
&0.0086$\pm$0.0006
&0.0087$\pm$0.0009
&0.0115$\pm$0.0007
&\underline{35.7794$\pm$0.1638}
&0.0035$\pm$0.0006
\\ 
\midrule

GraphMVP
&\textbf{0.5374$\pm$0.0093}
&3.0750$\pm$0.0057
&0.0062$\pm$0.0002
&0.0091$\pm$0.0007
&0.0102$\pm$0.0039
&85.8648$\pm$0.4281
&0.0029$\pm$0.0005

\\
GraphMVP-G
&0.5672$\pm$0.0086
&2.7505$\pm$0.1224
&0.0063$\pm$0.0001
&0.0074$\pm$0.0008
&0.0105$\pm$0.0019
&90.1874$\pm$4.9913
&0.0052$\pm$0.0002
\\
GraphMVP-C
&0.5578$\pm$0.0082
&2.8668$\pm$0.1915
&0.0061$\pm$0.0004
&0.0075$\pm$0.0006
&0.0103$\pm$0.0017
&78.5071$\pm$3.2950
&0.0034$\pm$0.0005
\\
Holi-Mol
&0.6441$\pm$0.0086
&4.4998$\pm$0.2511
&0.0095$\pm$00001
&0.0103$\pm$0.0009
&0.0151$\pm$0.0076
&71.7118$\pm$4.8325
&\underline{0.0028$\pm$0.0005}

\\
3D-InfoMax
&0.6447$\pm$0.0051
&5.9405$\pm$0.7233
&0.0117$\pm$0.0042
&0.0073$\pm$0.0009
&0.0142$\pm$0.0099
&77.1154$\pm$3.9725
&0.0077$\pm$0.0005

\\

\midrule
MVCIB (Ours) 
&\underline{0.5379$\pm$0.0063}
&\textbf{0.5196$\pm$2.7362}
&\textbf{0.0042$\pm$0.0001}
&\textbf{0.0044$\pm$0.0000}
&\textbf{0.0048$\pm$0.0001}
&\textbf{24.8314$\pm$1.3826}
&\textbf{0.0019$\pm$0.0000}
\\
\bottomrule
\end{tabular}
\label{tab:D3tasks}
\end{adjustbox}
\end{table*}

\begin{table*}[tb]
\caption{An interpretability comparison on functional group detection tasks in terms of $Fidelity-$/$+$.}
\centering
\renewcommand*{\arraystretch}{0.80}
\setlength{\tabcolsep}{5  pt}
\fontsize{9 pt}{9 pt}\selectfont
\begin{tabular}{l cc cc cc}
\toprule
\multirow{2}{*}{Methods}
&\multicolumn{2}{c}{BENZENE} 
&\multicolumn{2}{c}{Alkane Carbonyl} 
&\multicolumn{2}{c}{Fluoride Carbonyl} 
\\
\cmidrule(lr){2-3}
\cmidrule(lr){4-5}
\cmidrule(lr){6-7}
 
&$Fidelity-\downarrow$
&$Fidelity+\uparrow$
&$Fidelity-\downarrow$
&$Fidelity+\uparrow$
&$Fidelity-\downarrow$
&$Fidelity+\uparrow$
\\
\midrule











Infomax 

&0.363$\pm$0.041
&0.410$\pm$0.082

&0.353$\pm$0.021
&0.376$\pm$0.054

&0.331$\pm$0.032
&0.453$\pm$0.012
\\
JOAO

&{0.047$\pm$0.005}
&0.481$\pm$0.007

&0.263$\pm$0.005
&0.568$\pm$0.008

&0.183$\pm$0.007
&0.295$\pm$0.004
\\



GraphCL

&0.120$\pm$0.005
&0.469$\pm$0.008

&0.430$\pm$0.027
&0.578$\pm$0.016

&0.284$\pm$0.002
&0.570$\pm$0.001
\\

GraphLoG

&0.137$\pm$0.001
&0.459$\pm$0.007 

&0.355$\pm$0.002 
&0.695$\pm$0.007

&0.358$\pm$0.004 
&0.475$\pm$0.006
\\

\midrule

GraphFP

&0.093$\pm$0.006 
&{0.494$\pm$0.012}

&0.276$\pm$0.022 
&0.529$\pm$0.005

&0.263$\pm$0.026 
&0.595$\pm$0.034
\\
MICRO-Graph

&0.140$\pm$0.012 
&0.483$\pm$0.003 

&{0.155$\pm$0.037}
&0.524$\pm$0.015

&0.281$\pm$0.007 
&{0.595$\pm$0.002}
\\





GROVE

&0.064$\pm$0.029 
&0.489$\pm$0.010 

&0.183$\pm$0.041 
&0.518$\pm$0.007

&0.321$\pm$0.067 
&{0.628$\pm$0.038}
\\



S-CGIB

&{0.049$\pm$0.001}
&\textbf{{0.720$\pm$0.003}}	

&\underline{0.134$\pm$0.001}
&\underline{0.727$\pm$0.003}	

&0.133$\pm$0.002
&\underline{0.672$\pm$0.004}	
\\
\midrule

GraphMVP 
&0.105$\pm$0.014
&0.374$\pm$0.061
&0.157$\pm$0.003
&0.467$\pm$0.013	
&0.430$\pm$0.032
&0.484$\pm$0.037
\\

GraphMVP-G  
&0.179$\pm$0.003
&0.369$\pm$0.052
&0.195$\pm$0.004
&0.472$\pm$0.006	
&\underline{0.115$\pm$0.018}
&0.497$\pm$0.028
\\
GraphMVP-C 
&0.148$\pm$0.005
&0.485$\pm$0.011
&0.307$\pm$0.009	
&0.479$\pm$0.012
&0.255$\pm$0.052
&0.479$\pm$0.033
\\
Holi-Mol
&0.058$\pm$0.001
&0.477$\pm$0.016
&0.251$\pm$0.006	
&0.472$\pm$0.025
&0.194$\pm$0.011
&0.417$\pm$0.024
\\

3D-InfoMax
&\underline{0.037$\pm$0.005}
&0.492$\pm$0.012
&0.346$\pm$0.023
&0.466$\pm$0.004
&0.148$\pm$0.046
&0.513$\pm$0.041

\\

\midrule
MVCIB (Ours)
&\textbf{0.011$\pm$0.003}
&\underline{0.709$\pm$0.005}
&\textbf{0.118$\pm$0.005} 
&\textbf{0.752$\pm$0.003}
&\textbf{0.105$\pm$0.003}
&\textbf{0.708$\pm$0.002}
\\
\bottomrule
\end{tabular}
\label{tab:XGNNs}
\end{table*}

\begin{table*}[t]
\centering
\renewcommand*{\arraystretch}{0.80}
\caption{A Distinguishability (Dist.) and Alignment (Alig.) comparison in terms of Jensen-Shannon Divergence (JSD).}
\setlength{\tabcolsep}{3 pt}
\fontsize{9 pt}{9 pt}\selectfont
\begin{adjustbox}{width= \linewidth}
\begin{tabular}{l ccc ccc ccc}
\toprule
\multirow{2}{*}{Methods}
&\multicolumn{3}{c}{BENZENE} 
&\multicolumn{3}{c}{Alkane Carbonyl} 
&\multicolumn{3}{c}{Fluoride Carbonyl} 
\\
\cmidrule(lr){2-4}
\cmidrule(lr){5-7}
\cmidrule(lr){8-10}
&2D Dist.$\uparrow$
&3D Dist.$\uparrow$
&2D-3D Alig.$\downarrow$

&2D Dist.$\uparrow$
&3D Dist.$\uparrow$
&2D-3D Alig.$\downarrow$

&2D Dist.$\uparrow$
&3D Dist.$\uparrow$
&2D-3D Alig.$\downarrow$
\\ 
\midrule

GraphMVP
&0.113$\pm$0.002	
&0.048$\pm$0.000	
&0.007$\pm$0.001 

&0.093$\pm$0.001	
&0.065$\pm$0.001	
&0.025$\pm$0.001


&0.166$\pm$0.007	
&0.058$\pm$0.001
&0.016$\pm$0.002
\\ 
GraphMVP-G
&0.094$\pm$0.001	
&0.069$\pm$0.001
&0.002$\pm$0.000

&0.104$\pm$0.001
&0.052$\pm$0.001	
&0.023$\pm$0.002

&0.192$\pm$0.006	
&0.074$\pm$0.001
&0.013$\pm$0.001
\\
GraphMVP-C
&0.124$\pm$0.003
&0.055$\pm$0.001
&0.009$\pm$0.001

&0.094$\pm$0.001
&0.064$\pm$0.002
&\underline{0.016$\pm$0.002}


&0.183$\pm$0.006
&\underline{0.091$\pm$0.002}
&0.011$\pm$0.001
\\
Holi-Mol
&0.131$\pm$0.003	
&0.065$\pm$0.002	
&0.012$\pm$0.001

&0.112$\pm$0.002
&0.077$\pm$0.002
&0.021$\pm$0.001

&\underline{0.198$\pm$0.003}
&0.057$\pm$0.002
&0.014$\pm$0.001

\\ 
3D-InfoMax
&\underline{0.158$\pm$0.007}
&\underline{0.079$\pm$0.002}
&\underline{0.016$\pm$0.001}

&\underline{0.126$\pm$0.002}
&\underline{0.081$\pm$0.002}
&0.032$\pm$0.001

&0.184$\pm$0.003
&0.089$\pm$0.001
&\underline{0.008$\pm$0.000}

\\
\midrule
MVCIB (Ours)
&\textbf{0.185$\pm$0.006}
&\textbf{0.094$\pm$0.001}	
&\textbf{0.001$\pm$0.000}


&\textbf{0.156$\pm$0.002}	
&\textbf{0.108$\pm$0.001}	
&\textbf{0.011$\pm$0.002}

&\textbf{0.267$\pm$0.001}	
&\textbf{0.116$\pm$0.005}	
&\textbf{0.002$\pm$0.000}
\\
\bottomrule
\end{tabular}
\end{adjustbox}
\label{tab:well_distinction}
\end{table*}

\begin{table}[t]
\centering
\caption{
An expressiveness comparison in detecting non-isomorphic graph pairs in 1d- and 3d-WL datasets.
An ideal model should distinguish all pairs of non-isomorphic graphs.
}
\renewcommand*{\arraystretch}{0.80}
\setlength{\tabcolsep}{4  pt}
\fontsize{9 pt}{9 pt}\selectfont
\begin{tabular}{l  c  cccc c  }
\toprule
Types 
&\multicolumn{1}{c}{1d-WL}  
& \multicolumn{5}{c}{3d-WL} 
\\ 
\cmidrule(lr){2-2}
\cmidrule(lr){3-7}

Datasets
&GRAPH8C 
&SR1 
&SR2 
&SR3 
&SR4 
&SR5 
\\
 

$\#$ Graph pairs 
&61.8M 
&1 
&105 
&45
&6
&820
\\

\midrule
GIN 
&\underline{559}
&1
&105
&45
&6
&820
\\ 
\midrule
GraphCL
&9069
&1
&105
&45
&6
&\underline{532}

\\

GraphLoG
&7179
&1
&105
&45
&6
&820

\\
\midrule

MICRO-Graph
&21598	
&1
&105
&45
&6
&820
\\
MGSSL
&44102
&1
&105
&45
&6
&820
\\
S-CGIB
&1306
&1
&105
&45
&6
&820
\\
\midrule
GraphMVP
&1.4M
&1
&105
&45
&6
&820
\\
Holi-Mol
&1.2M
&1
&105	
&45
&6
&820
\\

3D-InfoMax
&1.1M
&1
&105
&45
&6
&820
\\


\midrule
MVCIB (Ours)
&\textbf{0}
&\textbf{0} 
&\textbf{0} 
&\textbf{0} 
&\textbf{0} 
&\textbf{0} 
\\ \hline
\end{tabular}
\label{tab:isomorphic_testing}
\end{table}

\begin{table}[t]
\centering
\caption{An ablation analysis on Subgraph Alignment (S.A.) and MVCIB modules.}
\setlength{\tabcolsep}{3 pt}
\begin{adjustbox}{width= \linewidth}
\begin{tabular}{l ccc ccc}
\toprule
\multirow{2}{*}{Methods}
&\multicolumn{3}{c}{ROC-AUC$\uparrow$}  
& \multicolumn{3}{c}{MAE$\downarrow$} 
\\
\cmidrule(lr){2-4}
\cmidrule(lr){5-7}
&BBBP
&BACE
&Tox21
&HOMO
&LUMO
&$\varepsilon_{gap}$
\\ 
\midrule


w/o S.A.
&{89.28}
&\underline{86.95}
&\underline{79.41}

&0.0067
&\underline{0.0075}
&\underline{0.0092}

\\

w/o MVCIB
&\underline{89.72}
&85.09
&78.92

&\underline{0.0061}
&0.0083
&0.0095



\\

\midrule
MVCIB (Ours)
&\textbf{90.76}
&\textbf{87.39}
&\textbf{81.03}

&\textbf{0.0042}
&\textbf{0.0044}
&\textbf{0.0048}


\\
\hline

\end{tabular}
\end{adjustbox}
\label{tab:abstudy_2}
\end{table}









Tab.~\ref{tab:acc1} shows the performance of MVCIB compared to baselines in terms of ROC-AUC on the MoleculeNet dataset.
We observed that:
(1) MVCIB consistently outperformed baselines on all datasets with a significant improvement.
For example, on the BBBP dataset, MVCIB achieved a 3.04\% improvement over the second-best model, further demonstrating its effectiveness.
(2) MVCIB performed well on multi-task learning datasets, e.g., Tox21 and SIDER, which consist of multiple classification tasks related to toxicity and drug side effect prediction. 
This indicates that MVCIB effectively captured representations that generalize well across diverse biochemical properties.
Moreover, we also demonstrated that MVCIB achieved a significant performance improvement over baselines on regression tasks (Appendix D.2).

\subsubsection{3D Geometric Property Regression Tasks}

Tab.~\ref{tab:D3tasks} presents the overall performance of MVCIB on seven 3D geometric properties from the QM9 quantum chemistry dataset.
We observed that MVCIB consistently outperformed the baselines, achieving the best results on six out of seven tasks, even surpassing models that incorporate additional 3D information, such as GraphMVP and Holo-Mol.
For instance, MVCIB achieved a 30.5\% improvement in $R^2$ for predicting electronic spatial extent, effectively capturing electron density within the 3D molecular structure.
Notably, MVCIB showed strong performance in predicting isotropic polarizability $\alpha$, HUMO, and LUMO predictions, showing its ability to learn geometry-aware representations that are essential for predicting quantum mechanical properties.

\subsection{Representation Quality Analysis}

\subsubsection{Interpretability Analysis}

There has been concern regarding the explainability of GNNs, which are black-box.
Thus, we conducted experiments to evaluate the quality of MVCIB's explanations by comparing them with ground-truth structural annotations \cite{agarwal2023evaluating}.
We selected the top 50\% of nodes with the highest cross-attention weights as the explanations.
We used \textit{Fidelity} metrics to evaluate how well the generated explanations align with MVCIB’s predictions, as shown in Tab.~\ref{tab:XGNNs}.
We observed that MVCIB consistently outperformed baselines, showing superior interpretability in detecting ground-truth substructures that determine the molecular properties.
In addition, qualitative analysis showed that substructures recognized by MVCIB are consistent with the knowledge from chemistry experts, such as the detection of alkane groups (Appendix D.3).


\subsubsection{Distinguishability and Alignment Analysis}

We conducted distinguishability and alignment analyses using Jensen-Shannon Divergence (JSD) on three datasets with functional group labels, as shown in Tab.~\ref{tab:well_distinction}.
In the distinguishability experiments (\textit{2D Dist.} and \textit{3D Dist.}), the goal is to assess whether molecular structures with different labels are distinguishable within each view, where a higher JSD indicates better discrimination.
In contrast, the alignment experiment (\textit{2D–3D Alig.}) evaluates whether molecular structures with the same label are aligned across views, where a lower JSD shows better alignment.
MVCIB achieved the highest JSD scores in the distinguishability tasks and the lowest JSD in the alignment task, showing its ability to discriminate between different molecular structures and consistently align representations across views.
In addition, we conducted cross-view reconstruction tasks to evaluate whether the shared information retains features from both molecular views (Appendix D.4).
The results showed that the shared representations preserved sufficient view-specific information to reconstruct each view, validating our assumption that all views originate from the same molecule.

\subsubsection{Expressiveness Analysis}


We evaluated the expressive power of MVCIB on isomorphism testing for 1d-WL and 3d-WL datasets.
A theoretical analysis of expressiveness is provided in Appendix A.2.
Tab.~\ref{tab:isomorphic_testing} shows the performance on Graph8c and five Strongly Regular Graphs (SRGs):
SR1 (16622),
SR2 (251256),
SR3 (261034),
SR4 (281264), 
and SR5 (291467).
We observed that MVCIB successfully distinguished both 1d- and 3d-WL isomorphism datasets, demonstrating its 3d-WL expressive power.
This implies that MVCIB effectively captured high-order substructures, which are critical for distinguishing different molecular structures.
In addition, MCVIB was capable of distinguishing isomers, which share similar 2D connectivity yet differ in 3D geometry and molecular properties (Appendix D.6).





\subsection{Model Analysis}

\subsubsection{Ablation Analysis}

We evaluated the contributions of the subgraph alignment and MVCIB modules to the overall performance of MVCIB, as shown in Tab.~\ref{tab:abstudy_2}.
The full results are given in Tab.~A4 in the Appendix.
This experiment showed that both modules jointly improved model performance, emphasizing their important roles in learning representations for effectively transferring knowledge to downstream tasks.
Notably, the MVCIB module showed a significant contribution, indicating its strong ability to capture shared information across molecular views.
We also observed that the 2D and 3D structural information are complementary, contributing jointly to the overall performance (Appendix D.5).

\subsubsection{Efficiency Analysis}

We demonstrated that MVCIB with pre-training converges remarkably faster than training from scratch, implying that it learned transferable and well-initialized representations (Appendix D.1).

\subsubsection{Sensitivity Analysis}

In the sensitivity analysis of subgraph size and interatomic distance, MVCIB achieved the optimal performance when the subgraph size is set to 3 and the interatomic distance threshold is 1.5~\text{\AA} (Appendix D.7).
Regarding the impact of GNN depth, we observed the best performance of MVCIB at $k=4$ layers (Appendix D.8).



\section{Conclusion}


We propose MVCIB, a GNN pre-training method, to improve multi-view learning of molecules by using the MVCIB principle and subgraph-level alignment.
Instead of maximizing the agreement between graph-level representations under contrastive objectives, MVCIB explicitly maximizes the shared information across views and minimizes the view-specific information.
Moreover, MVCIB could encode important substructures into representations by sampling and aligning them across views, enhancing the learning of chemical substructures during pre-training.
By doing so, MVCIB not only achieved superior predictive performance and interpretability but also achieved 3d-WL expressiveness power to distinguish different molecular structures.

 
\bibliography{aaai25}

\newpage



\section{Appendix}

\subsection{A. Proofs}
\label{app:proofs}

\subsection{A.1. Proof of Equation 15 in the main text}

Recall that given Equation 14 from the main text, our objective is to optimize the loss function $L_{MI}$, as:
\begin{eqnarray}
\label{eq:app_cib}
L_{MI} &= \left[  I\left( {G_{2D}};\textbf{Z}_{2D}^{CIB}|{G_{3D}} \right) + {I\left( {G_{3D}};\textbf{Z}_{3D}^{CIB}|{G_{2D}} \right)} \right]  \nonumber \\ 
& - \alpha  \left[  I\left( \textbf{Z}_{2D}^{CIB};{G_{3D}} \right) + I\left(G_{3D}^{CIB};{G_{2D}} \right)\right] . 
\end{eqnarray}

The first term can be minimized as:
\begin{align}
\label{eq:app_eq_1}
  & I\left( {G_{2D}};\textbf{Z}_{2D}^{CIB}|{G_{3D}} \right) \nonumber \\ 
  &= {\mathbb{E}_{{G_{2D}},\textbf{Z}_{2D}^{CIB},{G_{3D}}}}\left[ \text{log}\frac{{{p} }\left( \textbf{Z}_{2D}^{CIB}| {G_{2D}},{G_{3D}}\right)}{{{p} }\left( \textbf{Z}_{2D}^{CIB}| {G_{3D}} \right)} \right] \nonumber \\ 
 & ={\mathbb{E}_{{G_{2D}},\textbf{Z}_{2D}^{CIB},{G_{3D}}}}\left[ \text{log}\frac{{{p} }\left( \textbf{Z}_{2D}^{CIB}| {G_{2D}} \right) } {{{p}}\left( \textbf{Z}_{2D}^{CIB}|{G_{3D}} \right )} \right] \nonumber \\ 
  & ={\mathbb{E}_{{G_{2D}},\textbf{Z}_{2D}^{CIB},{G_{3D}}}}\left[ \text{log}\frac{{{p} }\left( \textbf{Z}_{2D}^{CIB}| {G_{2D}} \right) \ p\left( \textbf{Z}_{3D}^{CIB}| {G_{3D}} \right) }   {p\left( \textbf{Z}_{3D}^{CIB}|{G_{3D}}  \right ) \ p\left( \textbf{Z}_{2D}^{CIB}|{G_{3D}}  \right )} \right] \nonumber \\ 
 & ={{D}_{KL}}\left( {{p} }\left( \textbf{Z}_{2D}^{CIB}|{G_{2D}} \right)|| p \left( \textbf{Z}_{3D}^{CIB}|{G_{3D}} \right) \right) \nonumber  \\ 
 &  \ \ \ \ \ \  -{{D}_{KL}}\left( {{p}}\left( \textbf{Z}_{3D}^{CIB}|{G_{3D}} \right) p \left( \textbf{Z}_{2D}^{CIB}|{G_{3D}} \right) \right) \nonumber \\ 
 & \le {{D}_{KL}}\left( p\left( \textbf{Z}_{2D}^{CIB}|{G_{2D}} \right)||p\left( \textbf{Z}_{3D}^{CIB}|{G_{3D}} \right) \right) . \
\end{align}

Similarly, we can minimize the upper bound of the second term $I( {G_{3D}};\textbf{Z}_{3D}^{CIB}|{G_{2D}} )$, as:
\begin{align}
\label{eq:app_23}
I( {G_{3D}};\textbf{Z}_{3D}^{CIB}|{G_{2D}} ) \le 
{{D}_{KL}}( p( \textbf{Z}_{3D}^{CIB}|{G_{3D}} )||p( \textbf{Z}_{2D}^{CIB}|{G_{2D}} ) )  . \ 
\end{align}

For the third term in Equation \ref{eq:app_cib}, the mutual information $I( {{\textbf{Z}}_{2D}^{CIB}};{G_{3D}} )$ can be reformulated as:
\begin{align}
\label{eq:app_eq_2} 
   & I\left( {{\textbf{Z}}_{2D}^{CIB}};{G_{3D}} \right) \nonumber \\ 
   &=I\left( {{\textbf{Z}}_{2D}^{CIB}};{{\textbf{Z}}_{3D}^{CIB}}, {G_{3D}}\right)
  -I\left( {{\textbf{Z}}_{2D}^{CIB}};{{\textbf{Z}}_{3D}^{CIB}}|{G_{3D}} \right) \nonumber \\ 
 &=^* I\left( {{\textbf{Z}}_{2D}^{CIB}};{{\textbf{Z}}_{3D}^{CIB}},{G_{3D}} \right)  \nonumber \\
 &= I\left( {{\textbf{Z}}_{2D}^{CIB}};{{\textbf{Z}}_{2D}^{CIB}} \right)+I\left( {{\textbf{Z}}_{2D}^{CIB}};{{\textbf{Z}}_{2D}^{CIB}}|{G_{3D}} \right) \nonumber \\ 
 & \ge I\left( {{\textbf{Z}}_{2D}^{CIB}};{{\textbf{Z}}_{3D}^{CIB}} \right) \ ,  
\end{align}
where $*$ follows by assuming that $\textbf{Z}_{2D}^{CIB}$ and $\textbf{Z}_{3D}^{CIB}$ are conditionally independent given $G_{3D}$.

\begin{figure}[tb]
\centering
\includegraphics[width= \linewidth]{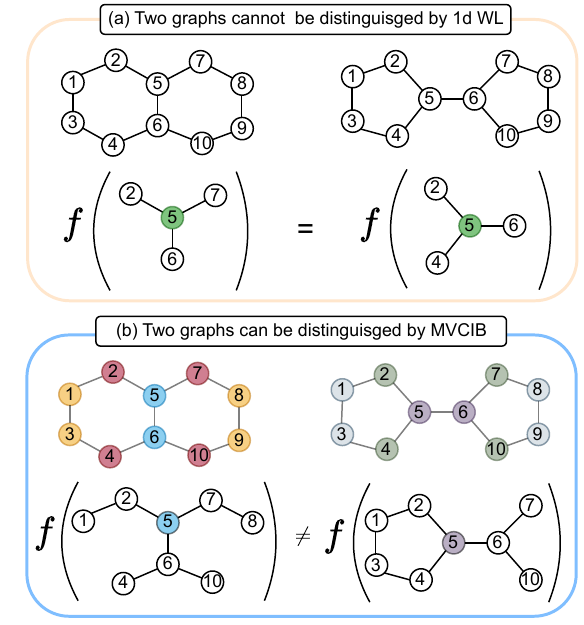}
\caption{
Given a target node \( v_5 \) and its surrounding substructures,  
(a) the 1d Weisfeiler–Lehman (1d-WL) test fails to distinguish the two non-isomorphic substructures rooted at \( v_5 \), even when the observation range is increased.  
(b) Our proposed model, MVCIB, successfully distinguishes the non-isomorphic substructures.
When the observation range is up to $2$-hop, the two substructures are different (the edges between nodes at the neighbours of node $v_5$) and can then be distinguished by our proposed model.
}
\label{fig:app_isomorphism}
\end{figure}

Similarly, the fourth term in Equation \ref{eq:app_cib}, i.e., $I( {{\textbf{Z}}_{3D}^{CIB}};{G_{2D}} )$, can be bounded as:
\begin{align}
\label{eq:app_4}
I( {{\textbf{Z}}_{3D}^{CIB}};{G_{2D}} ) \ge 
I( {{\textbf{Z}}_{3D}^{CIB}};{{\textbf{Z}}_{2D}^{CIB}} )\ .
\end{align}

By plugging Equations \ref{eq:app_eq_1}, \ref{eq:app_23}, \ref{eq:app_eq_2}, and \ref{eq:app_4} into Equation \ref{eq:app_cib}, we have:
\begin{align}
\label{eq:app_final}
L_{MI} & \le {{D}_{SKL}}\left( p\left( \textbf{Z}_{2D}^{CIB}|{G_{2D}} \right)||p\left( \textbf{Z}_{3D}^{CIB}|{G_{3D}} \right) \right) \nonumber \\ 
&  \ \ \ \ - \alpha I\left(  \textbf{Z}_{2D}^{CIB};  \textbf{Z}_{3D}^{CIB} \right) ,\ 
\end{align}
where $D_{{SKL}}( \cdot \,\|\, \cdot )$ denotes the symmetrized KL divergence between two distributions 
and $\alpha$ controls the trade-off between the two terms.
\subsection{A.2. Proof of Expressiveness Power of MVCIB}
\label{app:theorem}


Recall that our proposed model, MVCIB, is strictly more powerful than the 1-dimensional and 2-dimensional Weisfeiler–Lehman isomorphism tests (1d- and 2d-WL) in terms of expressiveness.

We first prove that if two graphs are identified as isomorphic by our proposed model, they are also recognized as isomorphic by the 1d-WL test.
Then, we present a pair of non-isomorphic graphs that can be distinguished by our proposed model but not by 1d-WL.
Together, these results imply that our proposed model is strictly more powerful than 1d-WL. 
Compared to the 2d-WL test, it can be concluded that 1d-WL and 2d-WL are equivalent in expressiveness for distinguishing graph isomorphism \cite{DBLP:conf/nips/MaronBSL19}.
In the proof, we use 1-hop ego-network subgraphs for our proposed model.

Assume that given two graphs $G$ and $H$ have the same number of nodes (otherwise easily determined as non-isomorphic) and are two non-isomorphic graphs, but our proposed model determines them as isomorphic.
Then for any iteration $t$, set $\{  \text{1d-WL}(G^{(t)} [N_1(v)]) | v \in V_{G} \}$ is the same as set $\{ \text{1d-WL}(H^{(t)} [N_1(v)]) | v \in V_{H} \}$.
Then there exists an ordering of nodes  $\{ v^G_1,v^G_2, \cdots, v^G_n \}$ in $G$ and $\{v^H_1,v^H_2, \cdots, v^H_n \}$ in $H$, such that for any node order $i = 1, \dots, N$, we have:
\begin{align}
\text{1d-WL}\left(G^{(t)}\left [N^G_1(v)\right]\right) = \text{1d-WL}\left(H^{(t)} \left[N^H_1(v) \right] \right) \ .
\end{align}
This implies that they are identical.
Thus, it can be deduced that $N^G_1(v)$ and $N^H_1(v)$ are hashed to the same label. 
Then for any iteration $t$ and any node with order, we have:
\begin{align}
\text{1d-WL}\left(N^G_1(v)\right) = \text{1d-WL}\left(N^H_1(v)\right)\ ,
\end{align}
where $N^G_1(v)$ and $N^H_1(v)$ denote the set of neighbours of node $v$.
This implies that 1d-WL also fails in distinguishing two non-isomorphic graphs $G$ and $H$.


In Fig.~\ref{fig:app_isomorphism}, two graphs are identified as isomorphic by the 1d-WL test, but accurately distinguished as non-isomorphic by MVCIB.
Specifically, given a target node \( v_5 \), the 1-dimensional Weisfeiler–Lehman (1d-WL) test fails to distinguish the two substructures rooted at \( v_5 \), even when the observation range is increased, as shown in Fig.~\ref{fig:app_isomorphism} (a).
In contrast, MVCIB successfully distinguished these non-isomorphic substructures based on distinctions in their surrounding substructures, as shown in Fig.~\ref{fig:app_isomorphism} (b).
This is because when the observation range is extended to the 2-hop neighborhood, CMVIB successfully distinguished the two substructures.
This result demonstrates that incorporating ego-networks enhances the expressive power of the 1d-WL test, enabling it to distinguish graph structures that 1d-WL fails to distinguish.

\subsection{B. Statistics of Datasets} 
\label{app:datasets}

We provide details on the datasets used in our experiments.
For the pre-training dataset, we used 456k unlabeled molecules from the ChEMBL database \cite{mayr2018large}.
For the 2D molecular downstream datasets, we utilized benchmark datasets from the MoleculeNet \cite{wu2018moleculenet}, including BBBP, Tox21, ToxCast, SIDER, ClinTox, MUV, ESOL, FreeSolv, and Lipo, mol-HIV, and BACE.
In addition, for evaluating performance on 3D geometric property regression, we adopted the QM9 dataset \cite{DBLP:conf/nips/HuFRNDL21}.
The detailed statistics for fine-tuning datasets are summarized in Tab.~\ref{tab:Benchmarks}.
For the datasets used in the interpretability analysis task, we used three datasets, including BENZENE, Alkane Carbonyl, and Fluoride Carbonyl, with their ground-truth explanations \cite{agarwal2023evaluating}, as:
\begin{itemize}
    \item  The BENZENE dataset consists of 12,000 molecules extracted from the ZINC dataset \cite{sterling2015zinc}, where each molecule is labeled with one of two classes based on the presence or absence of a benzene ring.
    \item  The Alkane Carbonyl dataset consists of 1,125 molecules, labeled based on the presence of an unbranched alkane and a carbonyl ($C=O$).
    \item  The Fluoride Carbonyl consists of 8,671 molecules, labeled according to the presence of fluoride ($F$-) and a carbonyl ($C=O$).
\end{itemize}

\subsection{C. Implementation Details}
\label{app:implement_detail}


\subsubsection{C.1. Model training}
The detailed hyperparameter settings are given in Tab.~\ref{tab:hyperparameter}.
We set the embedding dimension to 300.
The hyperparameters $\alpha$ and $\beta$ are determined with a grid search among $\{0.01, 0.1, 1.0, 10 \}$.
During the pre-training phase, we train the model for 500 epochs using the Adam optimizer with $1\times 10^{-4}$ learning rate and $1\times 10^{-5}$ weight decay.


\begin{table}[t]
\caption{Hyperparameters used in the experiments.}
\centering
\setlength{\tabcolsep}{3 pt}
\begin{tabular}{l c }
\toprule
Hyperparameters
&Values
\\ 
\midrule
Batch size
&64
\\
Number of Encoder layers
&4
\\
Embedding dimension
&300
\\
Number of pre-training epochs
&500
\\ 
Adam: initial learning rate
&$1\times 10^{-4}$
\\
Adam: weight decay
& $1\times 10^{-5}$
\\
Readout function
&SUM
\\
$\alpha$   
&1.0
\\
$\beta$   
&1.0
\\
\bottomrule
\end{tabular}
\label{tab:hyperparameter}
\end{table}

\begin{table*}[h]
\centering
\caption{The summary of statistics of datasets used in the experiments.}
\begin{tabular}{l c c c c  c   c}
\toprule
Category & Dataset&  \# Tasks & Task Type&  \# Graphs  &  Metric & Avg. \# nodes\\ 
\midrule

\multirow{2}{*}{Biophysics}
& mol-HIV 
& 1 
& Classification
& 41,127
& ROC-AUC
& 25.5
\\


& BACE
& 1
&Classification
&1,513 
&ROC-AUC
& 34.1
\\

\midrule
\multirow{6}{*}{Physiology}
&BBBP 
&1
&Classification 
&2,039 
&ROC-AUC
& 23.9
\\

& Tox21 
&12 
&Classification 
&7,831 
&ROC-AUC
& 18.6
\\

& ToxCast
&617 
&Classification 
&8,575 
&ROC-AUC
& 18.7
\\

& SIDER 
&27 
&Classification 
&1,427 
&ROC-AUC
& 33.6
\\

& ClinTox
&2
&Classification
&1,478 
&ROC-AUC
& 26.1
\\

& MUV
& 17
& Classification
&93,087
&ROC-AUC
& 24.2
\\


\midrule
\multirow{3}{*}{Physical Chemistry }
& Lipophilicity 
&1 
&Regression 
& 4,200
&RMSE
&  27.0
\\

& ESOL 
&1 
&Regression 
&1,128 
&RMSE
& 13.3
\\

& FreeSolv 
&1
&Regression
&642
&RMSE
& 8.7
\\
\midrule
Quantum Mechanics 
&QM9
&12 
&Regression 
&133,885 
& MAE
& 18.0
\\






\bottomrule
\end{tabular}
\label{tab:Benchmarks}
\end{table*}

\subsubsection{C.2. Baselines}

We considered three groups of baselines.
(1) The contrastive pretraining methods are 
Infomax \cite{velickovic2019deep}, 
JOAO \cite{DBLP:conf/icml/YouCSW21}, 
GraphCL \cite{DBLP:conf/nips/YouCSCWS20}, 
and GraphLoG \cite{DBLP:conf/icml/XuWNGT21}.
(2) The subgraph-based methods are 
MICRO-Graph \cite{DBLP:conf/aaai/Subramonian21}, 
MGSSL \cite{DBLP:conf/nips/ZhangLWLL21}, 
GraphFP \cite{DBLP:conf/nips/LuongS23}, 
GROVE \cite{DBLP:conf/nips/RongBXX0HH20}, 
and S-CGIB \cite{Hoang_Lee_2025}.
(3) The multi-view methods are
GraphMVP \cite{DBLP:conf/iclr/LiuWLLGT22},
Holi-Mol \cite{DBLP:journals/tmlr/KimNKLAS24},
3D-InfoMax \cite{DBLP:conf/icml/StarkBCTDGL22}.

We compare the proposed model against three categories of baselines: (i) contrastive learning methods, (ii) subgraph-based pre-training strategies, and (iii) multi-view pre-training strategies.
These baseline methods represent recent SOTA models in self-supervised pre-training for molecular graphs.
For fairness, we closely follow the experimental settings reported in the studies.

\textbf{First}, for contrastive learning strategies, we evaluated MVCIB against four methods:
\begin{itemize}
    \item  {Infomax} \cite{velickovic2019deep} strategy aims to maximize the agreement between the graph-level representations and their sampled subgraphs.
    \item  {JOAO} \cite{DBLP:conf/icml/YouCSW21} strategy generates a set of augmentation schemes, i.e., node dropping, subgraph augmentation, edge perturbation, feature masking, and identical, and tries to automatically find the useful augmentations.
 
    \item  {GraphCL} \cite{DBLP:conf/nips/YouCSCWS20} aims at generating multiple views of graphs based on four graph augmentations, i.e., node dropping, edge perturbation, node feature masking, and sampled subgraph, and then maximizes the agreement between these views based on contrastive objectives.
    \item  {GraphLoG} \cite{DBLP:conf/icml/XuWNGT21} discover the global graph structures by using hierarchical prototypes by contrasting graph pairs in a sampled batch.
\end{itemize}

\textbf{Second}, for subgraph-level pre-training strategies, we considered five methods:
\begin{itemize}
    \item  {GraphFP} \cite{DBLP:conf/nips/LuongS23} strategy aims to decompose input molecules into a set of subgraphs based on a dictionary (a bag of subgraphs based on subgraph frequent mining), which benefits the model in capturing semantic subgraphs.
    \item  {MICRO-Graph} \cite{DBLP:conf/aaai/Subramonian21} strategy aims to generate a bag of prototypical motifs and automatically learn the important functional group-like motifs.
    \item  {MGSSL} \cite{DBLP:conf/nips/ZhangLWLL21} strategy fragments the input molecules into a bag of functional groups based on an improved algorithm of BRICS by considering ring structures.
    \item  {GROVE} \cite{DBLP:conf/nips/RongBXX0HH20} extract 85 functional groups based on discovering frequent subgraphs based on RDKit and utilize a graph transformer architecture.
 
    \item  {S-CGIB} \cite{Hoang_Lee_2025} recognizes the core substructures and functional groups under the subgraph-conditioned graph information bottleneck principle.
\end{itemize}

\textbf{Third}, for multi-view pre-training strategies, we considered three methods:
\begin{itemize}

\item GraphMVP \cite{DBLP:conf/iclr/LiuWLLGT22} enhances a 2D molecular graph encoder by integrating rich and discriminative 3D geometric information.
There exist two variants of GraphMVP, i.e.,  GraphMVP-C and GraphMVP-G, which present different contrastive and generative Self-Supervised Learning objectives, respectively.

\item {HoliMol} \cite{DBLP:journals/tmlr/KimNKLAS24} adopts a fragmentation-based approach that decomposes molecules into chemically meaningful substructures, such as functional groups. 
In addition, HoliMol integrates the 3D geometric view as a complementary view to enrich contrastive learning, capturing both structural and spatial information to enhance molecular representations.

\item 3D-InfoMax \cite{DBLP:conf/icml/StarkBCTDGL22} aims to maximize the mutual information between 3D molecular representations and 2D molecular representations. This enables the GNN encoder to implicitly capture and encode 3D geometric information during the pre-training phase.

\end{itemize}

\begin{figure}[t]
\centering
\includegraphics[width=  \linewidth]{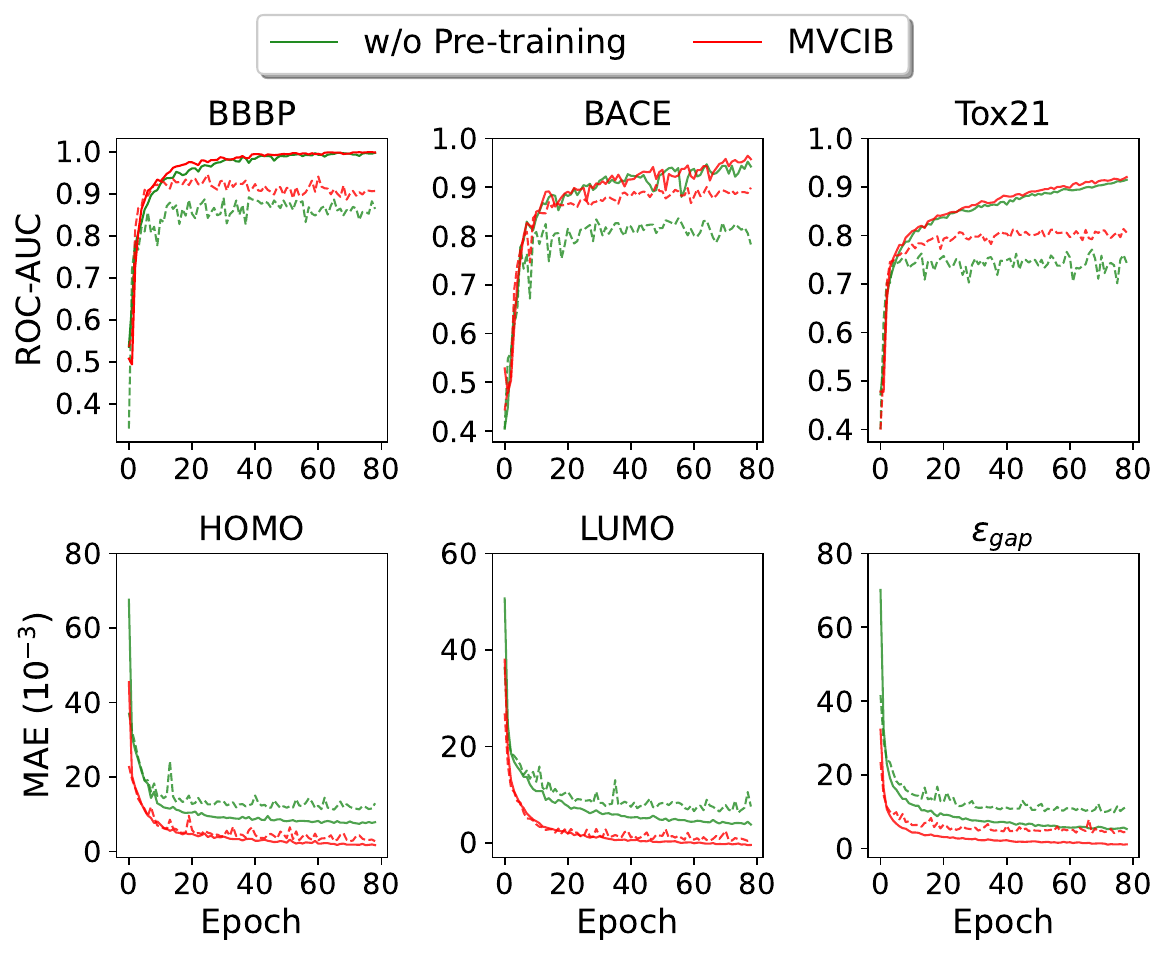}
\caption{An efficiency analysis for variants of our proposed model: MVCIB with and without pre-training.
The solid and dashed lines are training and validation curves, respectively.}
\label{fig:efficiency}
\end{figure}

\begin{figure*}[tb]
\centering
  \includegraphics[width=  \linewidth]{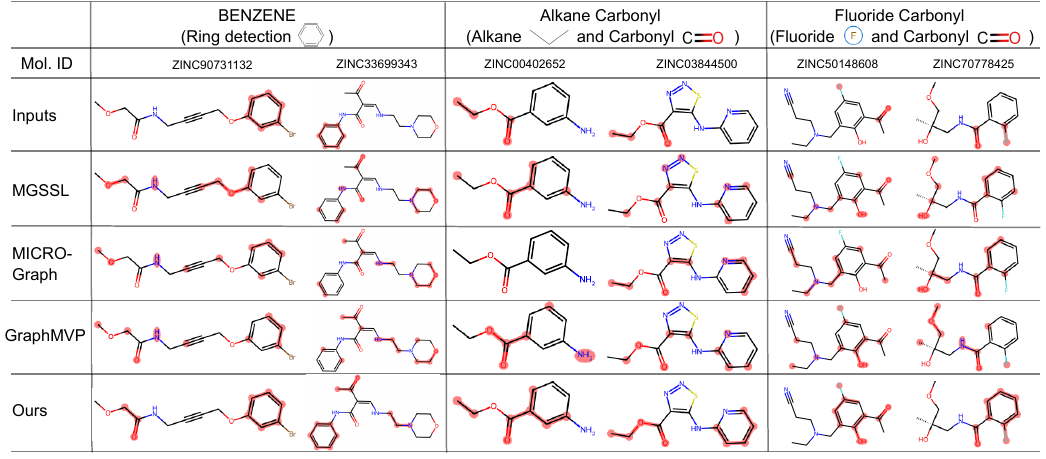}
  \caption{Qualitative analysis on functional group detection tasks.}
  \label{fig:qa}
\end{figure*}

\subsubsection{C.3. Training Resources} 

All experiments were conducted on two servers, each equipped with four NVIDIA RTX A5000 GPUs (24GB RAM/GPU).
Our proposed model was implemented and evaluated in Python 3.8.8, using both the PyTorch Geometric framework \cite{FeyLenssen2019} and the Deep Graph Library (DGL) \cite{wang2019dgl}.
All experiments were performed in an Ubuntu 20.04 LTS environment.

%
%

\subsection{D. Additional Experiments}

\subsubsection{D.1. Efficiency Analysis}
\label{app:add_effi}


We further evaluated the effects of our proposed model on convergence and generalization capability.
As shown in Fig.~\ref{fig:efficiency}, MVCIB consistently exhibited faster convergence compared to its counterpart trained from scratch, across both molecular property classification (upper) and 3D geometric property regression (lower) datasets.
This indicates that the pre-trained model effectively encoded chemical substructures, such as functional groups and local substructures, which serve as strong initialization signals for fine-tuning the models on downstream datasets.
Specifically, MVCIB reached optimal performance in significantly fewer training epochs, highlighting its ability to transfer learned information into robust downstream representations.
Moreover, the validation curves for MVCIB showed softer and more stable lines, indicating better generalization over unseen molecular data.

\subsubsection{D.2. Performance on Regression Tasks}
\label{app:reg_task}

We further conducted experiments on three datasets, e.g., FreeSolv, ESOL, and Lipophilicity, in terms of regression tasks, as presented in Tab.~\ref{tab:regression_task}.
We observed that:
(1) MVCIB outperformed existing multi-view methods across all three regression datasets, consistently achieving higher predictive accuracy.
This indicates that MVCIB more effectively captured chemically meaningful substructures across views, which is essential for predicting molecular regression properties such as solubility and lipophilicity.
(2) MVCIB exhibited significant advantages in datasets with a limited amount of labeled data available for training the model.
For example, MVCIB achieved an average performance improvement of 9.92\% on the ESOL dataset, which comprises only 1,128 labeled molecules.
This indicates that MVCIB learned the transferable representations that generalize well and adapt quickly to downstream datasets with limited labeled molecules.

\begin{table}[tb]
\caption{A performance comparison on molecular property regression tasks on the MoleculeNet dataset in terms of RMSE.
The top two are highlighted by \textbf{first} and \underline{second}.
}
\fontsize{9 pt}{9 pt}\selectfont
\setlength{\tabcolsep}{ 4.5 pt }
\centering
\begin{tabular}{l ccc}
\toprule
Methods
&FreeSolv 
&ESOL
&Lipophilicity
\\
\midrule
Infomax 
&3.033$\pm$0.026 
&2.953$\pm$0.049
&0.970$\pm$0.023
\\
JOAO 
&3.282$\pm$0.002 
&1.978$\pm$0.029
&1.093$\pm$0.097
\\
GraphCL 
&3.166$\pm$0.027
&1.390$\pm$0.363 
&1.014$\pm$0.018
\\
GraphLoG 
&2.335$\pm$0.052 
&1.542$\pm$0.026
&0.932$\pm$0.052
\\

\midrule
GraphFP
&2.528$\pm$0.016 
&2.136$\pm$0.096
&1.371$\pm$0.058
\\
MICRO-Graph
&{1.865$\pm$0.061} 
&{0.842$\pm$0.055}
&0.851$\pm$0.073
\\
MGSSL 
&2.940$\pm$0.051 
&2.936$\pm$0.071
&1.106$\pm$0.077
\\
GROVE
&2.712$\pm$0.327 
&1.237$\pm$0.403
&0.823$\pm$0.027
\\
S-CGIB
&{1.648$\pm$0.074} 
&\underline{0.816$\pm$0.019}
&\underline{0.762$\pm$0.042}
\\

\midrule

GraphMVP
&1.129$\pm$0.537
&0.913$\pm$0.006
&0.969$\pm$0.058

\\
GraphMVP-G
&\underline{1.068$\pm$0.172}
&0.902$\pm$0.027
&0.915$\pm$0.082

\\
GraphMVP-C
&1.082$\pm$0.051
&0.894$\pm$0.025
&0.892$\pm$0.073
\\
Holi-Mol
&2.189$\pm$0.077
&1.061$\pm$0.092 
&1.182$\pm$0.072
\\
3D-InfoMax
&2.217$\pm$0.126
&0.836$\pm$0.031
&0.915$\pm$0.089

\\

\midrule

MVCIB (Ours)
&\textbf{1.037$\pm$0.029}
&\textbf{0.735$\pm$0.026}
&\textbf{0.741$\pm$0.085}
\\

\bottomrule
\end{tabular}
\label{tab:regression_task}
\end{table}

\begin{table*}[t]
\centering
\caption{An ablation analysis on Subgraph-level Alignment (S.A.) and MVCIB module.}
\begin{tabular}{l ccc ccc}
\toprule
\multirow{2}{*}{Methods}
&\multicolumn{3}{c}{Classification Tasks (ROC-AUC$\uparrow$)}  
& \multicolumn{3}{c}{3D Geometry Regression Tasks (MAE $\downarrow$)} 
\\
\cmidrule(lr){2-4}
\cmidrule(lr){5-7}
&BBBP
&BACE
&Tox21
&HOMO
&LUMO
&$\varepsilon_{gap}$
\\ 
\midrule

w/o S.A.
&{89.28$\pm$0.51}
&\underline{86.95$\pm$0.58}
&\underline{79.41$\pm$0.39}
&0.0067$\pm$0.00003
&\underline{0.0075$\pm$0.00002}
&\underline{0.0092$\pm$0.00003}
\\

w/o MVCIB

&\underline{89.72$\pm$0.31}
&85.09$\pm$0.94
&78.92$\pm$0.66
&\underline{0.0061$\pm$0.00007}
&0.0083$\pm$0.00003
&0.0095$\pm$0.00008

\\

\midrule
MVCIB (Ours)
&\textbf{90.76$\pm$0.45}
&\textbf{87.39$\pm$0.21}
&\textbf{81.03$\pm$0.47}
&\textbf{0.0042$\pm$0.00001}
&\textbf{0.0044$\pm$0.00001}
&\textbf{0.0048$\pm$0.00002}
\\
\bottomrule

\end{tabular}

\label{tab:abstudy_2}
\end{table*}

\begin{table*}[t]
\caption{A performance comparison on cross-view reconstruction tasks in terms of MSE.}
\centering
\begin{tabular}{l ccc c}
\toprule

Methods


&BBBP
&BACE
&Tox21
&QM9

\\ 
\midrule

GraphMVP
&0.118$\pm$0.003
&0.148$\pm$0.009
&0.218$\pm$0.013

&0.341$\pm$0.013

\\
GraphMVP-G
&\underline{0.091$\pm$0.008}
&0.131$\pm$0.013
&0.174$\pm$0.019

&0.371$\pm$0.012
\\

GraphMVP-C
&0.127$\pm$0.016
&\underline{0.110$\pm$0.010}
&0.165$\pm$0.019

&0.235$\pm$0.012
\\

Holi-Mol
&0.096$\pm$0.005
&0.123$\pm$0.011
&0.180$\pm$0.102

&0.354$\pm$0.009
\\

3D-InfoMax
&0.119$\pm$0.017
&0.143$\pm$0.010
&\underline{0.142$\pm$0.012}

&\underline{0.247$\pm$0.021}
\\

\midrule

MVCIB (Ours)
&\textbf{0.031$\pm$0.003}
&\textbf{0.028$\pm$0.007}
&\textbf{0.094$\pm$0.009}

&\textbf{0.121$\pm$0.005}

\\ 
\bottomrule
\end{tabular}
\label{tab:cross_view2d}
\end{table*}

\begin{table*}[t]
\centering
\caption{An ablation study on single- and multi-view cases.}
\begin{tabular}{l ccc ccc}
\toprule
\multirow{2}{*}{Methods}
&\multicolumn{3}{c}{Classification Tasks (ROC-AUC$\uparrow$)}
& \multicolumn{3}{c}{3D Geometry Regression Tasks (MAE $\downarrow$)}
\\
\cmidrule(lr){2-4}
\cmidrule(lr){5-7}
&BBBP
&BACE
&Tox21 
&HOMO
&LUMO
&$\varepsilon_{gap}$
\\ 
\midrule
2D Only
&\underline{88.51$\pm$0.37}
&\underline{86.67$\pm$0.18}
&\underline{79.95$\pm$0.62}

&\underline{0.0072$\pm$0.00004} 
&0.0097$\pm$0.00007 
&0.0066$\pm$0.00013

\\
3D Only
&87.11$\pm$0.43
&84.72$\pm$0.32
&77.75$\pm$0.87
&0.0099$\pm$0.00008 
&\underline{0.0062$\pm$0.00012}
&\underline{0.0052$\pm$0.00003} 
\\



\midrule
MVCIB (Ours)
&\textbf{90.76$\pm$0.45}
&\textbf{87.39$\pm$0.21}
&\textbf{81.03$\pm$0.47}
&\textbf{0.0042$\pm$0.00011}
&\textbf{0.0044$\pm$0.00003} 
&\textbf{0.0048$\pm$0.00013}
\\
\bottomrule

\end{tabular}

\label{tab:abstudy}
\end{table*}

\subsubsection{D.3. Qualitative Analysis}
\label{app:qa}

Since functional group extraction could act as a bridge between pre-training and downstream molecular property prediction tasks, we are interested in its potential to provide domain-specific interpretability.
To qualitatively validate the interpretability of our proposed model, we visualize molecular substructures identified by MVCIB across three datasets: Benzene, Alkane Carbonyl, and Fluoride Carbonyl, as shown in Fig.~\ref{fig:qa}.
We observed that MVCIB delivered more accurate and chemically meaningful interpretations compared to baseline methods.
It implies that MCVIB could capture and align task-relevant substructures, such as functional groups, during the pre-training phase. 
By leveraging a subgraph-level alignment strategy guided by cross-attention mechanisms, the substructures highlighted by MVCIB aligned well with chemical knowledge from experts.
Specifically, the model consistently identified important substructures such as benzene rings, alkane chains, and carbonyl groups.
These results indicated that beyond improving model transferability, the interpretability offered by MVCIB significantly enhanced trust and practical usability in real-world applications such as drug discovery.


\subsubsection{D.4. Cross-View Reconstruction Tasks}
\label{app:ab_rec}

Motivated by our assumption that different molecular views originate from the same underlying molecule and thus contain shared information across views. 
Therefore, we conducted ablation studies to validate this hypothesis via cross-view reconstruction tasks on both 2D and 3D molecular datasets, as shown in Table~\ref{tab:cross_view2d}.
Specifically, for 2D molecular datasets, we employed the Mean Squared Error (MSE) to assess the model’s ability to reconstruct the original adjacency matrix from 3D molecular representations.
In contrast, for the 3D dataset, we aimed to reconstruct the interatomic distance based on only the representations from the 2D molecular view.
We observed that MVCIB significantly outperformed existing multi-view baselines in reconstructing both 2D topology and 3D geometry structures.
Unlike baselines that rely on contrastive learning objectives to maximize mutual information between graph-level representations, MVCIB explicitly focuses on discovering shared information and aligning important substructures across views.
This implies the importance of shared information discovery and subgraph-level alignment in learning the correct shared substructures across molecular views.
The robust performance in both reconstruction tasks delivered empirical evidence supporting our hypothesis that different molecular views encode complementary yet coherent information derived from the same underlying molecule.

\begin{table}[t]
\centering
\caption{
An expressiveness comparison in distinguishing isomers.
An ideal model should be able to distinguish every isomeric graph pair, as each isomer has different 3D geometry and molecular properties, even though they share the same 2D structures.
}
\renewcommand*{\arraystretch}{0.80}
\begin{tabular}{l  c c }
\toprule
Isomer type 
&\multicolumn{1}{c}{Cis/Trans isomers}  
& \multicolumn{1}{c}{Enantiomers}
\\ 
\cmidrule(lr){2-2}
\cmidrule(lr){3-3}

 


$\#$ Isomer pairs 
&500
&500
\\

\midrule
GIN 
&500
&500
\\ 
\midrule

JOAO 
&500
&500
\\

GraphCL
&500
&500
\\

GraphLoG
&500
&500
\\
\midrule

MICRO-Graph
&500
&500
\\
MGSSL
&500
&500
\\

S-CGIB
&500
&500
\\
\midrule
GraphMVP
&112
&500
\\

Holi-Mol
&\underline{95}
&171
\\

3D-InfoMax
&106
&\underline{129}
\\
\midrule
MVCIB (Ours)
&\textbf{0} 
&\textbf{0}
\\ \hline
\end{tabular}
\label{tab:isomorphic_testing}
\end{table}

\label{app:alignment}

\subsubsection{D.5. Ablation Study on Single-View \& Multi-View}

We evaluated the model performance under both multi-view and single-view settings, as summarized in Tab.~\ref{tab:abstudy}.
That is, we directly used the 2D molecular representations (\textit{2D Only}) or the 3D molecular representations (\textit{3D Only}) as the inputs for property predictions.
We observed that jointly leveraging both 2D and 3D molecular representations significantly enhanced predictive performance compared to utilizing either view individually.
This implies the complementary effect between learned representations from the two views, where 2D molecular features represent the explicit bond structures and 3D molecular features capture 3D geometric information that is essential for inferring intermolecular interactions.
Notably, MVCIB trained exclusively on the 3D view underperformed relative to both the 2D Only and multi-view settings.
This performance gap can be attributed to the absence of explicit bond topology in the 3D molecular representations, which can hinder the model’s ability to learn the correct chemical patterns.



\subsubsection{D.6. Expressiveness Analysis on Distinguishing Isomers}
\label{app:isomers}

To evaluate whether our proposed model can capture 3D geometric differences that significantly influence chemical properties, we conducted a study to assess the model’s ability to distinguish isomers, especially on Cis/Trans isomers and Enantiomers, as shown in Tab.~\ref{tab:isomorphic_testing}.
Cis/Trans isomers (geometric isomers) arise due to restricted rotation around a double bond or within a ring structure \cite{clayden2012organic}. 
They are classified based on the relative positions of two similar or high-priority groups: i.e., cis if the groups are on the same side and trans if they are on opposite sides.
In contrast, enantiomers are pairs of stereoisomers that are non-superimposable mirror images of one another. 
These arise when a molecule contains at least one chiral center, resulting in two stereoisomers with identical connectivity but different spatial information.

To the extent of our knowledge, there is no publicly available benchmark molecular dataset designed to evaluate the expressive power of GNNs in distinguishing isomers.
To address this gap, we introduce a new isomer recognition benchmark dataset specifically designed to evaluate GNN performance in distinguishing isomers, such as Cis/Trans isomers and Enantiomers.
Specifically, we randomly sampled isomeric molecules from the ZINC15 benchmark database \cite{sterling2015zinc}. 
For each molecule, we then generate its corresponding isomeric counterpart through expert-guided structural modifications, carefully preserving the smile string while modifying 3D geometric information.
The final dataset consists of 500 pairs of Cis/Trans isomers and 500 pairs of Enantiomers.
We will release the dataset publicly through our repository to support future research.
We observed that:
(1) MVCIB consistently distinguished between pairs of isomers, showing the model’s expressiveness in capturing isomeric variations.
This implies that MVCIB could capture high-order substructures, e.g., functional groups, and 3D geometric information, which empower MVCIB to recognize 3D geometry differences between isomers.
(2) Existing multi-view methods exhibited the ability to differentiate isomers, showing the importance of learning 3D geometric features in molecular learning.
However, these models overlooked the identification and alignment of chemical substructures across views, which limits their ability to distinguish isomeric structures.
(3) Single-view methods failed to distinguish between isomers due to the absence of explicit 3D geometry information.
To sum up, MVCIB showed an expressiveness in capturing and aligning high-order chemical substructures across views, enabling it to distinguish isomers.


\begin{figure}[t]
\centering 

\begin{subfigure}{0.47\textwidth}
\centering 
\includegraphics[width=\linewidth]{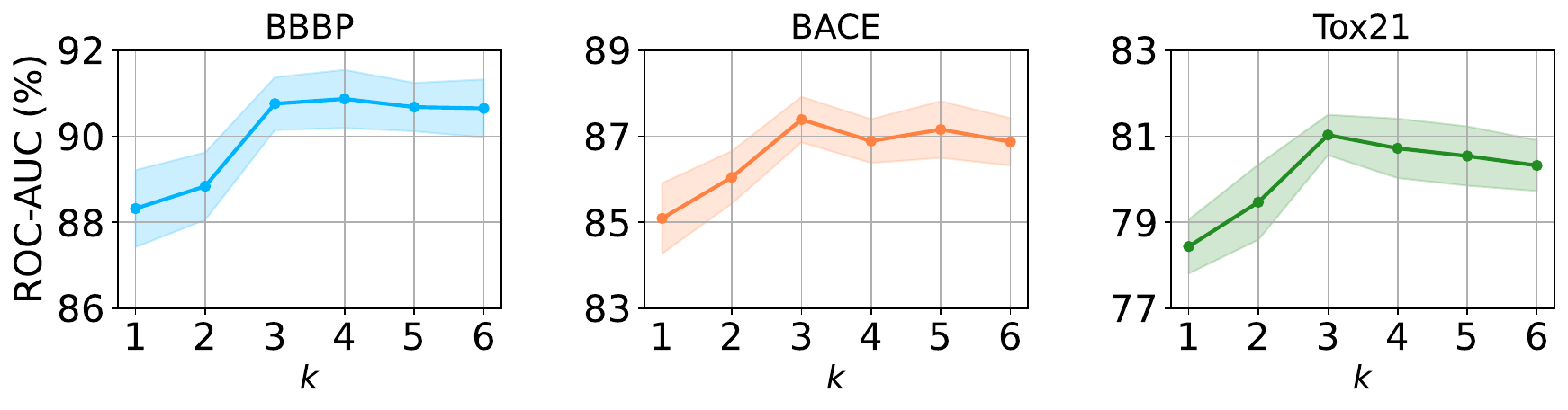}
\caption{On the $k$-hop radius.}
\label{fig:Lap_RWPE_a}
\end{subfigure}

\begin{subfigure}{0.47\textwidth}
\centering 
\includegraphics[width=\linewidth]{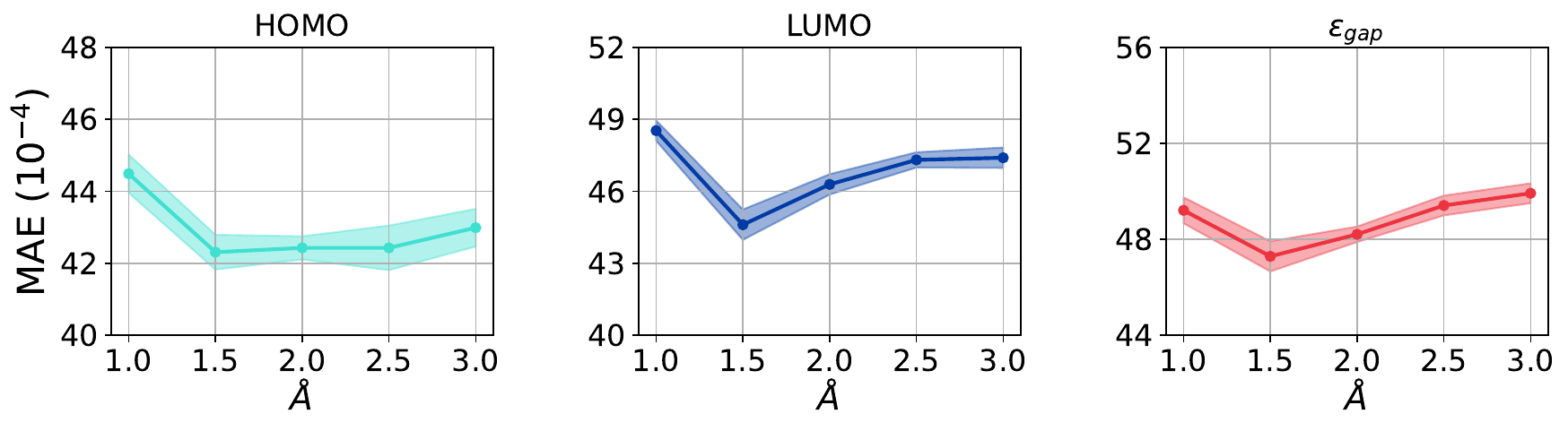}
\caption{On the interatomic distance cut-off ($\textup{\AA}$).}
\label{fig:Lap_RWPE_b}
\end{subfigure}\hfil 

\caption{A sensitivity analysis on subgraph sizes: (\textbf{a}) $k$-hop radius at 2D view and (\textbf{b}) distance cut-off at 3D view.}
\label{fig:subgraph_size_distance}
\end{figure}

\subsubsection{D.7. Sensitivity Analysis on Subgraph Sizes}


We further evaluated the performance of our proposed model according to the choice of subgraph size $k$ and interatomic distance cutoff, as shown in Fig.~\ref{fig:subgraph_size_distance}.
We observed that when $k$ was set to relatively small values, i.e., 3 and 4, our proposed model gained the optimal performance in almost all the datasets.
For example, in the BBBP dataset, MVCIB achieved the highest performance when the size $k=3$, corresponding to a 3-hop subgraph rooted at each node.
This implies that the subgraph size could be large enough to capture sufficient and significant subgraphs but not larger, so as to avoid noisy nodes.

Regarding the interatomic distance cutoff, we observed that MCVIB achieved the optimal performance when the cutoff was set to 1.5$\textup{\AA}$.
This indicates that a proper cutoff enabled MVCIB to capture the 3D spatial relationships between atoms without noise from irrelevant distant nodes.
That is, a small cutoff could overlook the important 3D geometric dependencies, while a too large cutoff value could introduce irrelevant information and degrade the model performance.

\begin{figure}[tb]
\centering
  \includegraphics[width=  \linewidth]{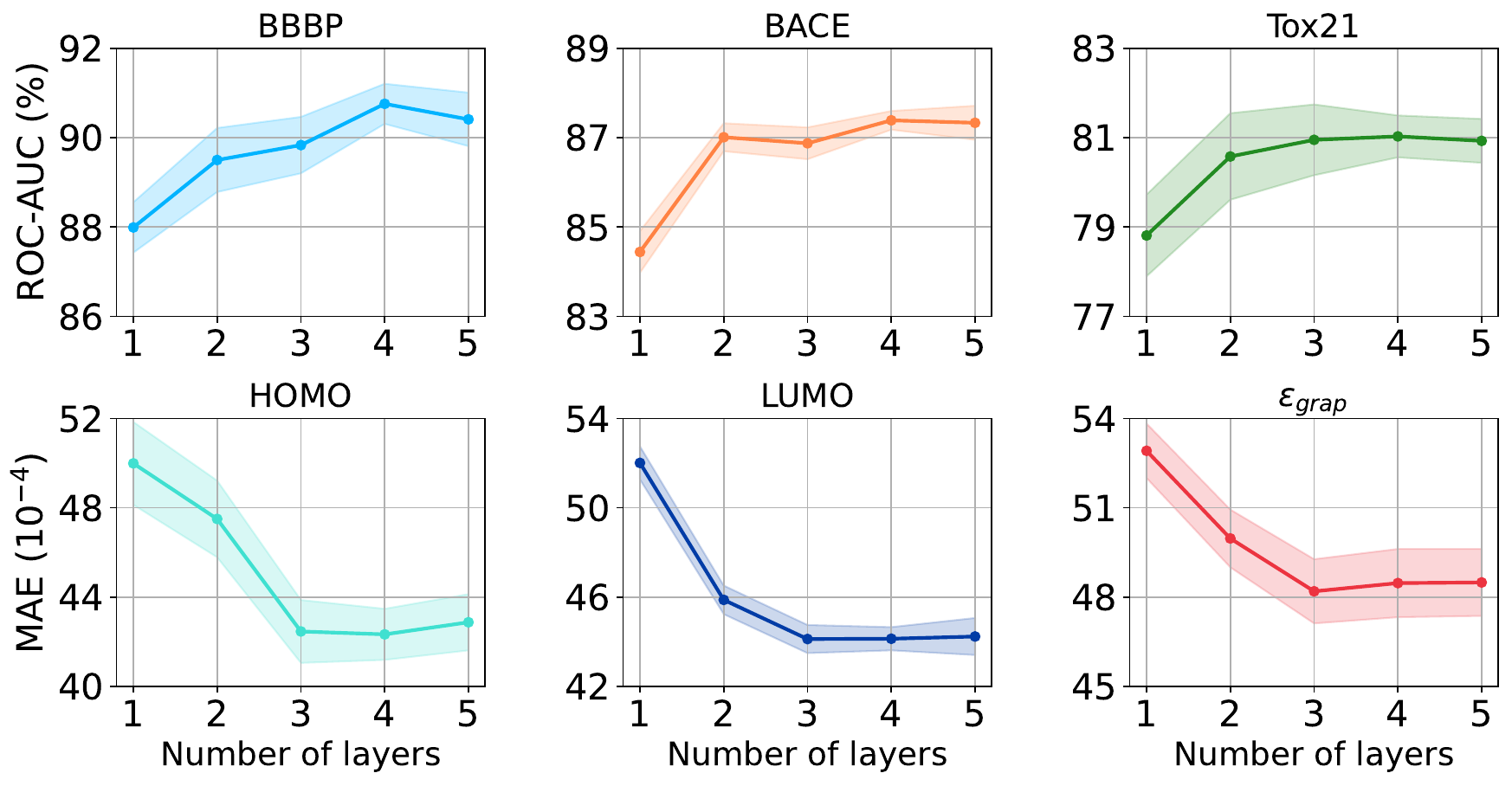}
  \caption{A sensitivity analysis on the number of GNN layers.}
  \label{fig:k_layer}
\end{figure}

\subsubsection{D.8. Sensitivity Analysis on the Number of Layers}

We further validated the performance according to changes in the number of GNN layers, as shown in Fig.~\ref{fig:k_layer}.
We observed that increasing the number of layers, especially to three or four, resulted in significant performance gains compared to other settings.
This suggests that a sufficiently deep GNN could enable the model to capture global structural information, which is essential for encoding high-order chemical substructures into the learned representations.
However, when the GNN depth continued to increase, the model performance started to degrade.
This is because the noisy and irrelevant information could hinder learning correct patterns, which are critical for distinguishing important molecular substructures.


\end{document}